\documentclass[a4paper,conference]{style/IEEEtran}
\IEEEoverridecommandlockouts
\usepackage{cite}
\usepackage{amsmath,amssymb,amsfonts}
\usepackage{algorithmic}
\usepackage[a4paper, left=37pt,right=37pt,top=52pt,bottom=116pt]{geometry}
\usepackage{graphicx}
\usepackage{textcomp}
\usepackage{xcolor}
\usepackage{multirow}

\usepackage{subfig}

\usepackage{pgfplots}
\usepackage{filecontents}

\usepackage[binary-units=true]{siunitx}
\DeclareSIUnit\px{px}
\DeclareSIUnit{\nothing}{\relax}

\usepackage{pifont}
	\newcommand{\cmark}{\ding{51}}%
	\newcommand{\xmark}{\ding{55}}%

\usepackage{hyperref}
\usepackage{cleveref}

\usepackage{threeparttable}

\usepackage{balance}

\definecolor{misc_object}{RGB}{128,   0, 128}
\definecolor{ego_vehicle}{RGB}{220, 220, 220}
\definecolor{car}{RGB}{0,  0,  142}
\definecolor{person}{RGB}{220, 20,   60}
\definecolor{bus}{RGB}{0,  60, 100}
\definecolor{forest}{RGB}{34, 139,  34}
\definecolor{bush}{RGB}{60, 179, 113}
\definecolor{sand}{RGB}{244, 164,  96}
\definecolor{asphalt}{RGB}{192, 192, 192}
\definecolor{gravel}{RGB}{105, 105, 105}
\definecolor{parking}{RGB}{250, 170, 160}
\definecolor{tree_crown}{RGB}{0, 220, 124}
\definecolor{tree_trunk}{RGB}{139,  69,  19}
\definecolor{soil}{RGB}{160,  82,  45}
\definecolor{animal}{RGB}{255, 182, 193}
\definecolor{building}{RGB}{1,  51,  73}
\definecolor{wall}{RGB}{0, 132, 111}
\definecolor{fence}{RGB}{190, 153, 153}
\definecolor{pole}{RGB}{153, 153, 153}
\definecolor{traffic_sign}{RGB}{220, 220,   0}
\definecolor{low_grass}{RGB}{154, 205,  50}
\definecolor{high_grass}{RGB}{0, 128,   0}
\definecolor{misc_vegetation}{RGB}{60, 179, 113}
\definecolor{sky}{RGB}{70, 130, 180}

\newcommand\numdataset{503} 					
\newcommand\numtrain{347} 						
\newcommand\numvaltest{78} 						

\newcommand\numclasses{44} 						
\newcommand\evalclasses{23} 					
\newcommand\numgroups{nine} 					
\newcommand\numvegetationclasses{seven}
\newcommand\numterrainclasses{four}
\newcommand\numcondensedclasses{14}
\newcommand\numrecordings{five}
\newcommand\namedataset{TAS500}

\newcommand\trainepochs{150} 					
\newcommand\gpu{Titan RTX GPU} 					
\newcommand\cuda{CUDA v10.0} 					
\newcommand\cudnn{CuDNN v7.4} 					

\newcommand{\mucarthree}{\mbox{MuCAR-3}}

\newlength{\tempheight}
\newlength{\tempwidth}

\newcommand{\rowname}[1]
{\rotatebox{90}{\makebox[\tempheight][c]{#1}}}

\newcommand{\columnname}[1]
{\makebox[\tempwidth][c]{#1}}

\usepackage[absolute]{textpos}
\newcommand{\copyrightstatement}{
	\begin{textblock*}{17cm}(20mm,1mm)    
		\noindent
		\footnotesize
		\copyright 2020 IEEE. Personal use of this material is permitted. Permission from IEEE must be
		obtained for all other uses, in any current or future media, including
		reprinting/republishing this material for advertising or promotional purposes, creating new
		collective works, for resale or redistribution to servers or lists, or reuse of any copyrighted
		component of this work in other works.
	\end{textblock*}
}

\begin{document}
\copyrightstatement
\bstctlcite{bibcontrol_etal4}

\title{
	A Fine-Grained Dataset and its Efficient Semantic Segmentation for Unstructured Driving Scenarios\\
\thanks{The authors gratefully acknowledge funding by the Federal Office of Bundeswehr Equipment, Information Technology and In-Service Support (BAAINBw).}
}

\author{\IEEEauthorblockN{Kai A. Metzger, Peter Mortimer and Hans-Joachim Wuensche}
\IEEEauthorblockA{Institute for Autonomous Systems Technology \\
Bundeswehr University Munich\\
Email: kai.metzger@unibw.de }
}

\maketitle

\begin{abstract}
	Research in autonomous driving for unstructured environments suffers from a lack of semantically labeled datasets compared to its urban counterpart. 
Urban and unstructured outdoor environments are challenging due to the varying lighting and weather conditions during a day and across seasons. 
In this paper, we introduce {\namedataset}, a novel semantic segmentation dataset for autonomous driving in unstructured environments. 
{\namedataset} offers fine-grained vegetation and terrain classes to learn drivable surfaces and natural obstacles in outdoor scenes effectively. 
We evaluate the performance of modern semantic segmentation models with an additional focus on their efficiency. 
Our experiments demonstrate the advantages of fine-grained semantic classes to improve the overall prediction accuracy, especially along the class boundaries.
The dataset and pretrained model are available at \href{https://www.mucar3.de/icpr2020-tas500}{mucar3.de/icpr2020-tas500}.
\end{abstract}

\begin{IEEEkeywords}
vegetation dataset, semantic segmentation, deep learning, autonomous driving, efficient
\end{IEEEkeywords}


\section{Introduction}
\label{sec:introduction}
Semantic scene understanding is a key capability for autonomous robot navigation in real-world environments, but current research in autonomous driving focuses mainly on urban, suburban, and highway scenes.
These scenes are considered as structured environments.
In terms of their scene statistics, structured environments often provide more explicit 
object boundaries and contain objects with strong structural regularities. 
We are interested in robot navigation in unstructured environments, such as paths through forests and along fields.
Here the occurrence of many semi-transparent classes such as tree foliage and the subtle color difference between vegetation types pose a challenge during image processing.

Our institute {\em Autonomous Systems Technology} has developed an autonomous vehicle named {\mucarthree} (Munich Cognitive Autonomous Robot Car) with full drive-by-wire capabilities.
{\mucarthree} is equipped with multiple sensors, such as LiDAR, vision systems, and inertial sensors.
Our research is motivated by the current perceptual capabilities of {\mucarthree}.

\begin{figure}[!t]
	\centering
	\subfloat{{\includegraphics[width=0.4875\textwidth]{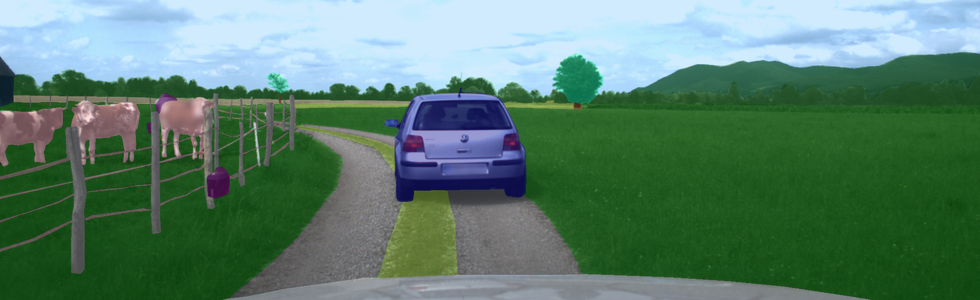} }}
	\\\vspace*{-0.8em}
	\subfloat{{\includegraphics[width=0.4875\textwidth]{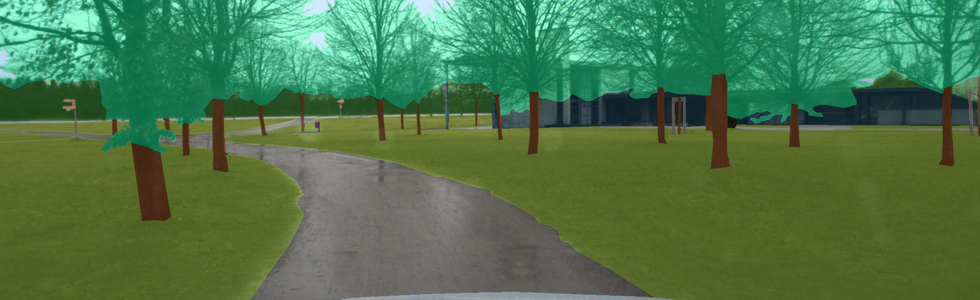} }}
	\\\vspace*{-0.8em}
	\subfloat{{\includegraphics[width=0.4875\textwidth]{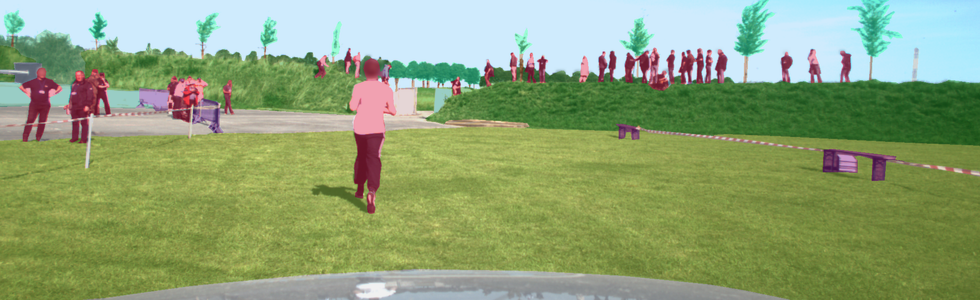} }}
	\caption{Example images from the TAS500 dataset. The TAS500 dataset contains fine-grained terrain and vegetation annotations of over 500 scenes in unstructured environments.}
	\label{fig:teaser}
	\vspace{-2mm}
\end{figure}
During test drives with {\mucarthree} in unstructured environments such as forest paths, it turned out that high grass or thin branches were perceived as an obstacle.
However, in other cases, there is impassible vegetation like barely visible tree trunks that {\mucarthree} should avoid during path planning.
Apart from a robust detection of passable and impassable vegetation, the domain of robot navigation requires a real-time capability from our semantic segmentation model to be able to react to changes in the environment. The semantic segmentation model should be able to process input images within a \SI{100}{\milli\second} cycle to keep our environment model updated at a frequency of at least \SI{10}{\hertz}.

In this paper, we study how a more fine-grained labeling policy affects the overall performance of the semantic segmentation in unstructured environments.
That is, we want to distinguish between drivable (grass) and non-drivable vegetation (\textit{bush} and \textit{tree trunk}), and detect different kinds of drivable surfaces (\textit{asphalt}, \textit{gravel}, \textit{soil}, \textit{sand}). \\
Our main contributions are as follows:
\begin{itemize}
	\item We release a novel dataset called {\namedataset} with more than \SI{500} RGB images with pixel-level ground truth annotations for driving in unstructured outdoor environments.
	\item We propose a training procedure that improves the vegetation and terrain segmentation performance in general and especially along the class boundaries using the fine-grained vegetation classes in our dataset. 
\end{itemize}
\vspace*{-0.8mm}
\begin{itemize}
	\item We compare state-of-the-art learning models for semantic segmentation using multiple quantitative measures with a focus on their run-time and memory efficiency for autonomous driving in unstructured outdoor environments (inference time below \SI{100}{\milli\second}).
\end{itemize}
%

\section{Related Work}
\label{sec:related_work}
This section provides a summary of datasets for autonomous robot applications, efficient semantic image segmentation methods and fine-grained image segmentation.

\subsection{Datasets}
 In the past decade, advances in autonomous robot technology and the successful use of learning algorithms in computer vision led to driving datasets gaining popularity.
Semantically annotated driving datasets are required to train learning algorithms and to provide a ground truth for quantitative comparisons.
Most datasets focus on urban driving environments \cite{brostow2009semantic, geiger2013vision, ros2016synthia}.

Cityscapes \cite{Cordts2016Cityscapes} and the KITTI semantic segmentation benchmark \cite{alhaija2018kittisemsegbenchmark} cover scenes in dense urban traffic and provide instance-level semantic segmentation for 30 classes.
Berkeley Deep Drive \cite{yu2018bdd100k} and A2D2 \cite{geyer2020a2d2} provide up to 40 object classes for semantic segmentation. 
These large datasets provide multi-granularity annotations for an increasing number of object classes but fail to capture the diversity of vegetation and terrain.
FreiburgForest \cite{Valada_2016_ISER} contains multi-modal images in forested environments and FieldSAFE \cite{kragh2017fieldsafe} in agriculture; both datasets provide pixel-wise ground truth annotations for six classes, including a semantic class representing vegetation. 
The 2016 Sugar Beets Dataset \cite{chebrolu2017agricultural} distinguishes between different object classes such as sugar beets and several weed species. 
The latter dataset provides images from different growth stages of vegetation but only includes scenes of agricultural fields. 

To the best of our knowledge, there is a lack of datasets that provide ground truth data for challenging computer vision tasks in unstructured environments with fine-grained distinctions between drivable vegetation, non-drivable vegetation, and multiple terrain types.

Our contributed dataset {\namedataset} provides fine-grained annotations for both terrain and vegetation. We subdivide drivable surfaces into {\numterrainclasses} terrain subclasses and vegetation into {\numvegetationclasses} different subclasses.

\subsection{Efficient Semantic Image Segmentation}
Probabilistic graphical models like Markov random fields \cite{kame:geman1984stochastic} and conditional random fields \cite{kame:kohli2008crf, kame:plath2009crf} have traditionally been used for pixel-wise image labeling.

In recent years, the success of neural networks for image classification \cite{kame:krizhevsky2012imagenet, kame:simonyan2015vgg} motivated the development of neural networks specifically for semantic image segmentation such as FCN \cite{kame:long2015fully}, SegNet \cite{kame:badrinarayanan2017segnet}, PSPNet \cite{kame:zhao2017pyramid}, and DeepLabv3 \cite{kame:chen2017rethinking}. 

Neural network architectures for semantic segmentation often face a trade-off between the output resolution, inference speed, memory usage during inference, and prediction accuracy.
For instance, very Deep Convolutional Neural Networks (DCNNs) such as DeepLabv3 \cite{kame:chen2017rethinking} and HRNetV2 \cite{kame:sun2019high} boosted the performance to $81.3 \%$ and $81.8 \%$ class mIoU (mean Intersection over Union), respectively, on the Cityscapes pixel-level segmentation task but are inefficient in terms of their inference speed.

FCN \cite{kame:long2015fully} introduced skip connections at different resolutions and SegNet \cite{kame:badrinarayanan2017segnet} applied this in an encoder-decoder framework.
Two-branch frameworks enabled efficient semantic segmentation compared to preceding deep architectures.
Modern approaches use multi-resolution branches, which can cope with high-resolution images and achieve inference times below \SI{100}{\milli\second} \cite{kame:paszke2016enet, kame:romera2017erfnet, kame:yu2018bisenet, kame:zhao2018icnet, kame:poudel2019fast}.

The necessity to classify multiple sensor data streams with limited computational resources and the real-time requirements in an autonomous vehicle motivate our comparative study of efficient semantic image segmentation methods.

\subsection{Fine-Grained Image Segmentation}

Limited research has been done in the field of fine-grained semantic image segmentation. In fine-grained visual categorization (FGVC), methods are developed to distinguish between subordinate object categories with low inter-class variance, such as classifying species of birds \cite{kame:ji2020fgvc, kame:zheng2017fgvc}. This task is often limited to image classification and does not extend to semantic segmentation. 

In the 3D domain advances have been made in object understanding when segmenting objects into their parts \cite{kame:kalogerakis2017shapepfcn, kame:zhao20193dpointcapsule}. In 2D, approaches based on object parts were used for fine-grained object understanding \cite{kame:liu2020partguidedediting, kame:lu2018partstates}. 

The recent MSeg \cite{kame:lambert2020mseg} is composed of multiple semantic segmentation datasets from varying domains, leading to a total of 194 categories. The MSeg dataset does not provide driving scenes from unstructured environments and lays its focus on performing robustly across many visual domains. 

In comparison, our dataset {\namedataset} provides fine-grained segmentations for the specific domain of robot navigation in unstructured environments. {\namedataset} distinguishes between subordinate object categories for classes with high visual and semantic similarity such as \textit{low grass} and \textit{high grass} while also segmenting objects into semantic parts such as a tree into its \textit{tree crown} and \textit{tree trunk}. 

\section{Dataset}
\label{sec:vegetation_dataset}
Our goal to enhance semantic scene understanding in unstructured environments led to the development of this dataset.
\subsection{Data Specifications}
The data was collected using the autonomous vehicle {\mucarthree} \cite{kame:himmelsbach2011autonomous}.
Our vision system is mounted on a multifocal active/reactive camera platform called MarVEye \cite{tas:unterholzner2010iv-hybrid-adaptive-control}.
A camera sensor provides color images at \SI{2.0}{\mega P} resolution.
We recorded data with a frame rate of \SI{10}{\hertz} and cut off most of the sky and ego vehicle hood from all images. 
The final images have a resolution of $620$ $\times$ \SI{2026}{\px}.
Our label rate amounts to around \SI{0.1}{\hertz}, and we consequently provide a pixel-wise semantic mask for every hundredth recorded image. \\
The dataset can be requested at \href{https://www.mucar3.de/icpr2020-tas500}{mucar3.de/icpr2020-tas500}.
\subsection{Semantic Classes and Annotations}
We provide fine-grained annotations at pixel level. 
Some of these classes were adopted from the Cityscapes dataset \cite{Cordts2016Cityscapes}, but we split up categories such as \textit{nature} and \textit{flat} to further subdivide vegetation 
as well as drivable surfaces. 
We annotated images from five test drives, as described in \Cref{fig:chp4:num-frames-duration}.
\begin{table}[bt]
\vspace{3mm}
\caption{Number of labeled frames and corresponding duration for the {\numrecordings} annotated sequences in {\namedataset}.}
\label{fig:chp4:num-frames-duration}
\begin{center}
	\begin{tabular}{c|c|c|c|c}
		Sequence
		& \begin{tabular}[c]{@{}c@{}}Season\end{tabular}
		& \begin{tabular}[c]{@{}c@{}}Weather\\condition\end{tabular}
		& \begin{tabular}[c]{@{}c@{}}Number of\\labeled frames\end{tabular}
		& \begin{tabular}[c]{@{}c@{}}Duration\\(mm:ss)\end{tabular}\\
		\hline\rule{0pt}{1.0\normalbaselineskip}
		\hspace{-0.94mm}1& Summer 
		& sunny & 99 	& 06:57 \\
		2 				 & Fall 
		& sunny & 57 	& 04:53 \\
		3 				 & Fall 
		& sunny & 129 	& 09:08	\\
		4 				 & Fall 
		& sunny & 126 	& 13:49 \\
		5 				 & Spring 
		& rainy & 92 	& 15:02 \\
		\hline		\rule{0pt}{1.0\normalbaselineskip}
		\hspace{-1.0mm}Total			& --
		& -- & {\numdataset}	& 49:49
	\end{tabular}
\end{center}
\end{table}
Our {\numdataset} pixel-level semantic masks were labeled with a self-developed
application, which works similar to the image annotation tool LabelMe \cite{kame:russell2008labelme}, but our tool is also capable of annotating 3D point clouds. 
We use closed polygonal chains to label objects and structures in camera images; exactly one class label was assigned for every pixel.
The annotation speed was increased by reusing the object boundaries and totaled approximately \SI{45}{\minute} per frame, including a quality control process.

Our labeling policy defines {\numclasses} class labels that are categorized into {\numgroups} groups: \textit{animal}, \textit{construction}, \textit{human}, \textit{object}, \textit{sky}, \textit{terrain}, \textit{vegetation}, \textit{vehicle}, and \textit{void}.
Moreover, we want to ensure compatibility with existing datasets \cite{brostow2009semantic, geiger2013vision, Cordts2016Cityscapes}.
We therefore include most classes from the Cityscapes dataset but exclude rare classes. Thus we only use the classes \textit{car} and \textit{bus} from the vehicle category.

For our experiments in this paper we map infrequent classes such as \textit{rider} to the closest corresponding category (here: \textit{person}), while too specific class distinctions such as the classes \textit{street light} and \textit{pole} are consolidated into one super-class (here: \textit{pole}).
Through this process we arrive at {\evalclasses} classes that we use for our experiments. 
The color coding for the classes and corresponding groups are shown in \Cref{fig:chp3:num-pixels}.

The labeling policy for {\namedataset} was designed for driving in unstructured environments.
Common objects were subdivided if they might require different planning behavior from an autonomous vehicle. Specifically vegetation classes can be split into drivable (e.g. \textit{high grass}) and non-drivable vegetation (e.g. \textit{bush}). Our fine-grained vegetation annotations allow the distinction between passable (e.g. \textit{tree crown}) and impassible (e.g. \textit{tree trunk}) classes within the non-drivable vegetation subcategory. Passable vegetation covers any type of vegetation that an autonomous vehicle could graze during path planning.

\subsection{Dataset Splits}
We split the semantically annotated images into separate training, validation, and test sets.
The images of the {\numrecordings} recordings were not randomly chosen as we wanted to ensure a balanced distribution from each labeled stream.
Our dataset is partitioned into a \SI{70}{\percent}-\SI{15}{\percent}-\SI{15}{\percent} train, validation, and test split.
The training set consists of {\numtrain} images; validation and test sets consist of {\numvaltest} images each.
\subsection{Statistical Analysis}
We compare {\namedataset} to other driving datasets in terms of the relative class distributions of the terrain and vegetation categories.
\Cref{fig:chp3:statistical-analysis} compares the urban driving datasets Cityscapes and KITTI with our dataset {\namedataset} and FreiburgForest, which were both recorded in unstructured environments.
Especially the terrain and vegetation categories are highly relevant for unstructured and forested scenarios.

Our dataset naturally has a high relative pixel occurrence for the \textit{terrain} as well as the drivable and non-drivable \textit{vegetation} classes compared to the other datasets. 
We provide fewer pixels for the group \textit{terrain} as Cityscapes and higher relative numbers as KITTI and FreiburgForest.
Besides, our dataset provides more vegetation pixels per image when compared to the driving datasets Cityscapes and KITTI and is on par with FreiburgForest
(last column, \Cref{fig:chp3:statistical-analysis}).
Moreover, our dataset incorporates varying weather conditions: the first four recordings cover sunny weather conditions and the last sequence rainy weather (\Cref{fig:chp4:num-frames-duration}). 
One data sequence each was recorded during spring and summer, respectively, and the remaining three recordings cover the fall season.
\begin{table}[bt]
	\vspace{3mm}
	\caption{Relative number of terrain and vegetation pixels per image for Cityscapes, KITTI, FreiburgForest, and our {\namedataset} dataset for the training sets.}
	\label{fig:chp3:statistical-analysis}
	\begin{center}
		\begin{threeparttable}
		\begin{tabular}{c|c|c|c}
			Dataset
			& \begin{tabular}[c]{@{}c@{}}Resolution ($H \times W$)\end{tabular}
			& \begin{tabular}[c]{@{}c@{}}Terrain ($\%$)\end{tabular}
			& \begin{tabular}[c]{@{}c@{}}Vegetation\tnote{\textdaggerdbl}~ ($\%$)\end{tabular} \\
			\hline
			\rule{0pt}{1.0\normalbaselineskip}
			Cityscapes\tnote{\textdagger}
			& $1024 \times 2048$ 
			& 38.0 			
			& 14.1 \\ 		
			KITTI\tnote{\textdagger}
			& $375 \times 1242$ 
			& 10.0 			
			& 30.3 \\ 		
			FreiburgForest
			& $487 \times 880$ 
			& 9.3 			
			& 25.7 + 39.9\\	
			{\namedataset} (ours)
			& $620 \times 2026$  
			& 18.4 			
			& 35.3 + 23.3 	
		\end{tabular}
		\begin{tablenotes}[para]
		\footnotesize
		\item[\textdagger] The terrain class is composed of the road and sidewalk classes.
		\item[\textdaggerdbl] Vegetation is split into drivable and non-drivable where applicable.
	\end{tablenotes}
	\end{threeparttable}
	\end{center}
\end{table}
\begin{figure*}[tbh]
	\begin{tikzpicture}
	\begin{axis}[
	every axis plot post/.style={/pgf/number format/fixed},
	width=1.0\textwidth,
	height=5cm,
	bar width=10pt,
	enlarge y limits=0.3,
	ymin=100000,
	ymax=1000000000,
	xmin=0.0, 
	xmax=23.0,
	xtick={0.2, 3.3, 8.6, 11.5, 13.7, 15.3, 18.0, 19.7, 21.5, 22.8},
	xticklabels={terrain,vegetation,construction,vehicle,sky,object,human,animal,void},
	x tick label as interval = true,
	restrict x to domain=0:20,
	ymode=log,
	ybar=3pt,
	ymajorgrids=true,
	ylabel={number of pixels},
	ylabel near ticks,
    nodes near coords,
	every node near coord/.append style={rotate=90, anchor=west},
	point meta=explicit symbolic,
	]
	\addplot[fill=asphalt,draw=none] coordinates {(7.7,44870171) [\scriptsize\textcolor{black}{asphalt}]};
	\addplot[fill=gravel,draw=none] coordinates {(7.7,25166793) [\scriptsize\textcolor{black}{gravel}]};
	\addplot[fill=soil,draw=none] coordinates {(7.7,3169035) [\scriptsize\textcolor{black}{soil}]};
	\addplot[fill=sand,draw=none] coordinates {(7.7,183702) [\scriptsize\textcolor{black}{sand$^*$}]};
	\addplot[fill=bush,draw=none] coordinates {(8.5,79861212) [\scriptsize\textcolor{black}{bush}]};
	\addplot[fill=forest,draw=none] coordinates {(8.5,72185405) [\scriptsize\textcolor{black}{forest}]};
	\addplot[fill=low_grass,draw=none] coordinates {(8.5,54085188) [\scriptsize\textcolor{black}{low grass}]};
	\addplot[fill=high_grass,draw=none] coordinates {(8.5,30076973) [\scriptsize\textcolor{black}{high grass}]};
	\addplot[fill=misc_vegetation,draw=none] coordinates {(8.5,5709992) [\scriptsize\textcolor{black}{misc. veg.}]};
	\addplot[fill=tree_crown,draw=none] coordinates {(8.5,5341737) [\scriptsize\textcolor{black}{tree crown}]};
	\addplot[fill=tree_trunk,draw=none] coordinates {(8.5,685644) [\scriptsize\textcolor{black}{tree trunk}]};
	\addplot[fill=building,draw=none] coordinates {(9.4,6701160) [\scriptsize\textcolor{black}{building}]};
	\addplot[fill=fence,draw=none] coordinates {(9.4,11869372) [\scriptsize\textcolor{black}{fence}]};
	\addplot[fill=wall,draw=none] coordinates {(9.4,1422834) [\scriptsize\textcolor{black}{wall$^*$}]};
	\addplot[fill=car,draw=none] coordinates {(10.4,4654762) [\scriptsize\textcolor{black}{car}]};
	\addplot[fill=bus,draw=none] coordinates {(10.4,320803) [\scriptsize\textcolor{black}{bus}]};
	\addplot[fill=sky,draw=none] coordinates {(11.3,44593592) [\scriptsize\textcolor{black}{sky}]};
	\addplot[fill=misc_object,draw=none] coordinates {(12.2,1712550) [\scriptsize\textcolor{black}{misc. object}]};
	\addplot[fill=pole,draw=none] coordinates {(12.2,423176) [\scriptsize\textcolor{black}{pole}]};
	\addplot[fill=traffic_sign,draw=none] coordinates {(12.2,84760) [\scriptsize\textcolor{black}{traffic sign$^*$}]};
	\addplot[fill=person,draw=none] coordinates {(13.2,791706) [\scriptsize\textcolor{black}{person}]};
	\addplot[fill=animal,draw=none] coordinates {(14.3,83556) [\scriptsize\textcolor{black}{animal$^*$}]};
	\addplot[fill=ego_vehicle,draw=none] coordinates {(15.2,19701974) [\scriptsize\textcolor{black}{ego vehicle}]};
	\end{axis}

	\node at (13,3.1) {$^*$excluded from evaluation};
	\end{tikzpicture}
	\caption{Number of fine-grained pixels (y-axis) per class and their associated category (x-axis).}
	\label{fig:chp3:num-pixels}
\end{figure*}
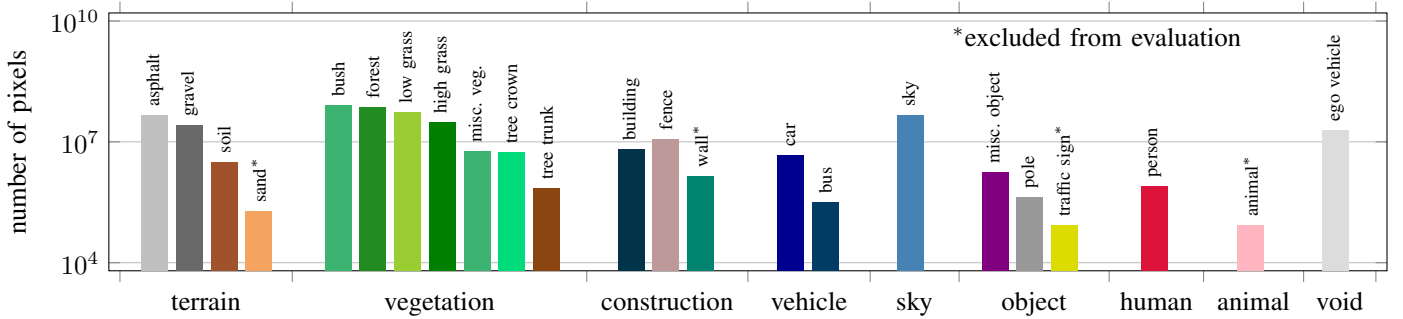

\section{Methodology} 
\label{sec:methodology}
In this section we compare different DCNN architectures and describe evaluation metrics which we use in our subsequent experiments.
\subsection{Architecture}
The trade-off between performance and run-time of DCNNs brings up the necessity of carefully choosing the input image size.
One can either choose to downsample high-resolution images (e.g., resize the image and interpolate between pixels) or utilize DCNNs that can process an arbitrary input size.

AdapNet was specifically designed to process images with a fixed input size ($384 \times 786$); PSPNet and OCNet \cite{kame:yuan2018ocnet} also process images with higher resolutions 
($769 \times 769$).
We prefer high-resolution segmentations to reliably detect thin obstacles like poles, which are only represented by very few pixels in lower resolutions.
AdapNet and Fast-SCNN share a similar performance ($\approx$ 70 mIoU) on the Cityscapes test set.
There are other state-of-the-art methods such as OCNet, which perform better ($\approx$ 80 mIoU) on the Cityscapes test set, but when applied to our autonomous driving scenario they do not fit our speed requirements (\textless~\SI{100}{\milli\second}). Such networks are therefore not considered in our experiments.

In our experiment we compare recent works such as ENet and Fast-SCNN. These efficient DCNNs incorporate up to two orders of magnitude fewer parameters than their predecessors PSPNet, SegNet, and DeepLabv3. They are therefore applicable in mobile robotics with limited hardware resources.

Additionally, Fast-SCNN encodes spatial information and global context in a learning-to-downsample module, allowing the network to process images of different resolutions.
We do not consider pretraining the networks on other datasets, since our network comparison includes DCNNs with low learning capacity (number of parameters $\lesssim$\ 1 million), which do not benefit from pretraining on big datasets or weakly labeled data \cite{kame:poudel2019fast}.

\subsection{Evaluation Metrics for Semantic Segmentation}
The Intersection over Union (IoU), also called Jaccard index, is a standard metric to evaluate object detection and semantic segmentation results. 
It is defined as the intersection of two areas divided by the size of their union.
The IoU is applicable for evaluating a semantic segmentation by comparing the prediction region with the ground truth region for a class as follows:  $\textnormal{IoU} =
\frac{\textnormal{TP}}{\textnormal{TP} + \textnormal{FN} + \textnormal{FP}}$. The IoU is a common metric for evaluating semantic image segmentations, but is known to be biased towards 
object instances that cover a large image area \cite{Cordts2016Cityscapes}. For the domain of autonomous driving this bias in an evaluation metric is unwanted, because it does not incentivize the use of models that can detect thin structures and smaller obstacles like poles and tree trunks.

To address this we use the Boundary Jaccard (BJ) score \cite{kame:fernandez2018new} as an additional evaluation metric.
An advantage of the BJ metric is that it penalizes wrong border predictions stronger than the IoU metric by considering both the global and contour accuracy.

In the definition of the $\textnormal{BJ}^c$ for a class $c$, the full binary image of class $c$ is denoted by $\textnormal{S}^{c}_{\text{ps}}$ in the prediction and by ${S}^{c}_{\text{gt}}$ in the ground truth. The pixels along the borders of the binary image are denoted by $\textnormal{B}^{c}_{\text{ps}}$ in the prediction and by $\textnormal{B}^{c}_{\text{gt}}$ in the ground truth.
We compute the distances from the boundary of the ground truth binary map $\textnormal{B}^{c}_{\text{gt}}$ to the binary class map of the prediction $\textnormal{S}^{c}_{\text{ps}}$ and vice versa.

The border threshold $\theta$ specifies the distance (in pixels) from the respective border we still consider as correctly classified, favoring matching predictions closer to the border:
\begin{equation}
\resizebox{.5 \textwidth}{!}
{$ \textnormal{TP}_{B_{\text{gt}}}^c \mkern-4mu=\mkern-8mu
	\sum\limits_{ 	x \in \textnormal{B}^{c}_{\text{gt}} } \mkern-5mu z \textnormal{ with } z = \mkern-1mu \begin{cases} 1 - (d(x,S^{c}_{\text{ps}}) / \theta)^2 \quad \mkern-5mu \textnormal{ if } d(x,\textnormal{S}^{c}_{\text{ps}}) < \theta\\ 0 \qquad\qquad\qquad\quad\;\;\;\text{otherwise} \end{cases}
	$}
\end{equation}
\begin{equation}
\resizebox{.5 \textwidth}{!}
{$ \textnormal{TP}_{B_{\text{ps}}}^c \mkern-4mu=\mkern-8mu
	\sum\limits_{ 	x \in \textnormal{B}^{c}_{\text{ps}} } \mkern-5mu z \textnormal{ with } z = \mkern-1mu \begin{cases} 1 - (d(x,S^{c}_{\text{gt}}) / \theta)^2 \;\;\textnormal{ if } d(x,\textnormal{S}^{c}_{\text{gt}}) < \theta\\ 0 \qquad\qquad\qquad\quad\;\;\;\text{otherwise} \end{cases}
	$}
\end{equation}

We adjust the border threshold $\theta$ in the BJ metric in accordance to the increasing image resolution. 
This should counterbalance the increased difficulty of predicting the exact class borders in higher-resolution images.
A threshold of $\theta = 5$ is used for the smallest resolution; for higher resolutions threshold values we chose $\theta = 8$ and $\theta = 12$. 
The total number of true positives is calculated as $\textnormal{TP}^c = \textnormal{TP}_{\textnormal{B}_{gt}}^c + \textnormal{TP}_{\textnormal{B}_{ps}}^c$.
Finally, the false negatives are computed as $\textnormal{FN}^c =  |\textnormal{B}_{gt}^c| - \textnormal{TP}_{\textnormal{B}_{gt}}^c$ and false positives as $\textnormal{FP}^c =  |\textnormal{B}_{ps}^c| - \textnormal{TP}_{\textnormal{B}_{ps}}^c$.
The BJ metric for class $c$ is calculated according to the Jaccard index equation:
\begin{equation}
\textnormal{BJ}^c = \frac{\textnormal{TP}^c}{\textnormal{TP}^c + \textnormal{FN}^c + \textnormal{FP}^c} \; ,
\end{equation}
The BJ metric is, similarly to the IoU metric, above zero if the predicted segmentation map and the ground truth map for a class have any overlap. 
Both the IoU and the BJ metric are used for evaluation in our experiments.


\section{Experiments}
\label{sec:experiments}
This section describes the training procedure and the experiments including the network selection, granularity level investigation, data augmentation and an ablation study.
\subsection{Training Procedure}
Our implementations use the open-source machine learning libraries PyTorch \cite{kame:paszke2019pytorch} and Tensorflow \cite{kame:martin2015tensorflow}.
All network implementations used in the experiments are integrated in open-source training suites \cite{nn:seif2018, kame:Efficient-Segmentation-Networks}.

We first compare promising efficient semantic segmentation models. At this stage we do not apply any data augmentation techniques yet and select a neural network for the subsequent experiments. 
In the next step we analyze if a subdivision of the standard label classes vegetation and terrain improve the prediction performance.
Finally, we apply general data augmentation techniques to our dataset and report this with the other improvements on our final model in an ablation study. 

We choose to resize our original image size to ($384 \times 768$) or ($512 \times 1024$), respectively, depending on the necessary input sizes for each model.
The batch size is set to $\text{bs} = 8$, we employ {\trainepochs} training epochs, use the Adam optimizer \cite{kame:kingma2014adam}, and train on a single {\gpu} with {\cuda} and {\cudnn}.
Following the previous works of \cite{kame:chen2017rethinking}, we employ a polynomial learning rate policy with an initial learning rate of $0.01$.
The learning rate is multiplied by $(1 - \frac{\text{epoch}}{\text{epoch}_{\text{max}}})^{0.9}$ for every iteration.
We employ a class-balanced cross entropy loss and do not apply any data augmentation methods.
Training processes are conducted for different DCNNs with a total of {\numtrain} images from the training set of our unstructured dataset if not stated otherwise.
We decided to not use pretraining in our training processes because DCNNs with low learning capacity do not benefit significantly from it \cite{kame:poudel2019fast}.

The segmentation model PSPNet uses the ResNet101 \cite{kame:he2016deep, kame:he2016identity} feature extraction model 
and is pretrained on ImageNet \cite{kame:deng2009imagenet}.
No class balancing is applied for the PSPNet and AdapNet architectures.
The other network architectures in the experiment do not need weight initialization.
\subsection{Comparison of Efficient Semantic Segmentation Models}
\label{cpt5:section-comparison}
%
For this experiment, we condense the classes of the categories terrain and vegetation from \Cref{fig:chp3:num-pixels} into one class each, resulting in {\numcondensedclasses} class labels.
\Cref{tab:experiments:cnn} reports the inference speed as well as 
the mean Intersection over Union (mIoU) 
and mean Boundary Jaccard (mBJ) for each model.
We argue that the low-resolutions models can not detect fine details adequately and, thus, result in a lower performance compared to high-resolution models. 
The BJ metric confirms this as it penalizes wrong boundary predictions: both low-resolution models do not perform well on the BJ metric.
Models that process images with higher resolutions perform better even when possessing a lower learning capacity in terms of the number of network parameters.
Therefore, high-resolution models incorporate better segmentation capabilities.
Low-capacity models like ENet and Fast-SCNN are highly efficient and use up to two orders of magnitude fewer parameters than PSPNet.

We choose the Fast-SCNN architecture because it yields a comparable performance as other semantic image segmentation models, while being the only model to be able to process the original resolution ($620 \times 2026$) as well as arbitrary image sizes.
Moreover, it has a low memory footprint during inference and provides a fast inference time.
\begin{table}[tb]
	\vspace{1.3mm}
	\centering
	\caption{Comparison of efficient semantic image segmentation methods with reference implementations on the validation set of our {\namedataset} dataset.}
	\label{tab:experiments:cnn}
	\begin{tabular}{@{}l|c|c|c|l@{}}
		Method
		& \begin{tabular}[c]{@{}c@{}}Memory\\footprint ($\si{\giga\byte}$)\end{tabular}
		& \begin{tabular}[c]{@{}c@{}}Runtime\\(\si{\milli\second})\end{tabular}
		& \begin{tabular}[c]{@{}c@{}}Val mIoU\\($\%$)\end{tabular}
		& \begin{tabular}[c]{@{}c@{}}Val mBJ\\($\%$)\end{tabular}\\
		\hline
		\rule{0pt}{1.0\normalbaselineskip}
		$\mathbf{384 \times 786~\si{\px}}$ & & & &$\mathbf{\theta = 5}$\\
		\hspace{4mm}PSPNet \cite{kame:zhao2017pyramid}
		& 5.2
		& 31 
		& 53.66
		& \hspace{2mm}13.46 \\ 
		\hspace{4mm}AdapNet \cite{kame:valada2017adapnet}
		& 4.9
		& 28 
		& 51.91
		& \hspace{3.35mm}9.72 \\ 
		%
		\hline\rule{0pt}{1.0\normalbaselineskip}
		$\mathbf{512 \times 1024~\si{\px}}$ & & & &$\mathbf{\theta = 8}$\\
		\hspace{4mm}SQNet \cite{kame:treml2016speeding}
		& 8.9
		& 23 
		& 57.64
		& \hspace{2mm}63.46 \\ 
		\hspace{4mm}ERFNet \cite{kame:romera2017erfnet}
		& 3.1
		& 17 
		& \textbf{59.42}
		& \hspace{2mm}61.39 \\ 
		\hspace{4mm}ENet \cite{kame:yu2018bisenet}
		& \textbf{1.7}
		& 28 
		& 51.35
		& \hspace{2mm}64.49 \\ 
		\hspace{4mm}LinkNet \cite{kame:chaurasia2017linknet}
		& 1.9
		& 11 
		& 57.04
		& \hspace{2mm}\textbf{66.34} \\ 
		\hline\rule{0pt}{1.0\normalbaselineskip}
		$\mathbf{620 \times 2026~\si{\px}}$&&&&$\mathbf{\theta = 12}$\\
		\hspace{4mm}Fast-SCNN \cite{kame:poudel2019fast}
		& \textbf{1.7}
		& \textbf{9} 
		& 53.00
		& \hspace{2mm}60.73 \\ 
	\end{tabular}
\end{table}
\subsection{Subdivision of the Vegetation Class}
Our autonomous vehicle {\mucarthree} continues on paths through unstructured terrain or forested environments and needs to distinguish between drivable and non-drivable vegetation.
The classification of fine-grained vegetation is a mandatory perceptual capability for {\mucarthree} to traverse a narrow forest path. This for example requires driving under \textit{tree crowns} but avoiding the \textit{tree trunks}.
In other scenarios there are \textit{bushes} that the robot could graze (e.g., push away thin branches or leaves) in order to continue on its route.
We investigate the further subdivision of the non-drivable vegetation class into passable (\textit{bush, tree crown}) and impassable (\textit{forest, miscellaneous vegetation, tree trunk}) vegetation classes to allow for more complex path planning for autonomous robots on forested paths.

\begin{figure*}[t]%
	\vspace{-2mm}
	\centering
	\subfloat{{\includegraphics[width=0.2477\textwidth]{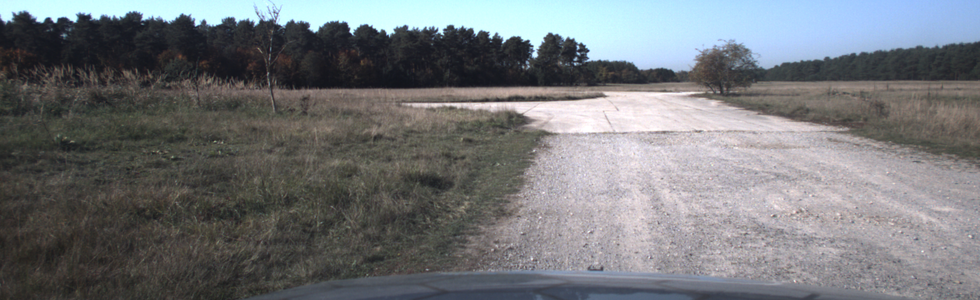} }}
	\hspace*{-0.45em}
	\subfloat{{\includegraphics[width=0.2477\textwidth]{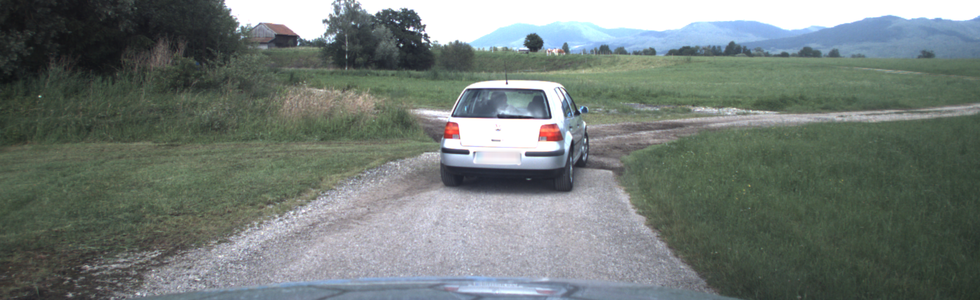} }}
	\hspace*{-0.45em}
	\subfloat{{\includegraphics[width=0.2477\textwidth]{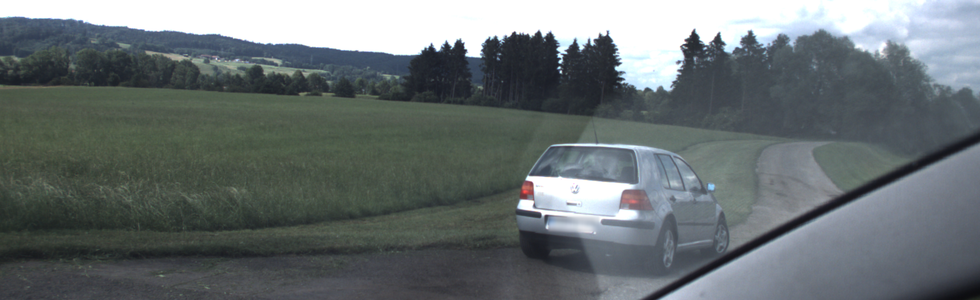} }}
	\hspace*{-0.45em}
	\subfloat{{\includegraphics[width=0.2477\textwidth]{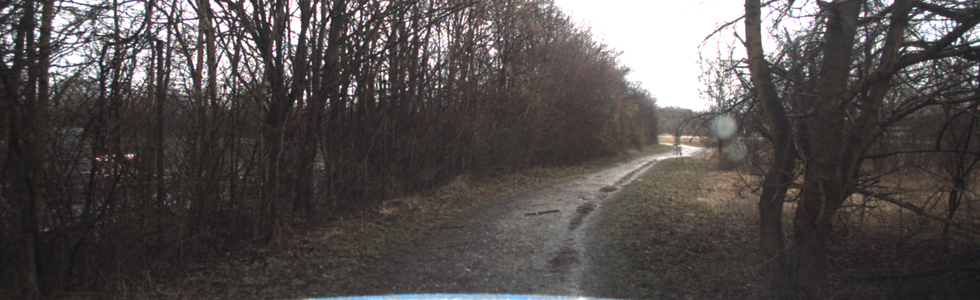} }}
	\\\vspace*{-0.8em}
	\subfloat{{\includegraphics[width=0.2477\textwidth]{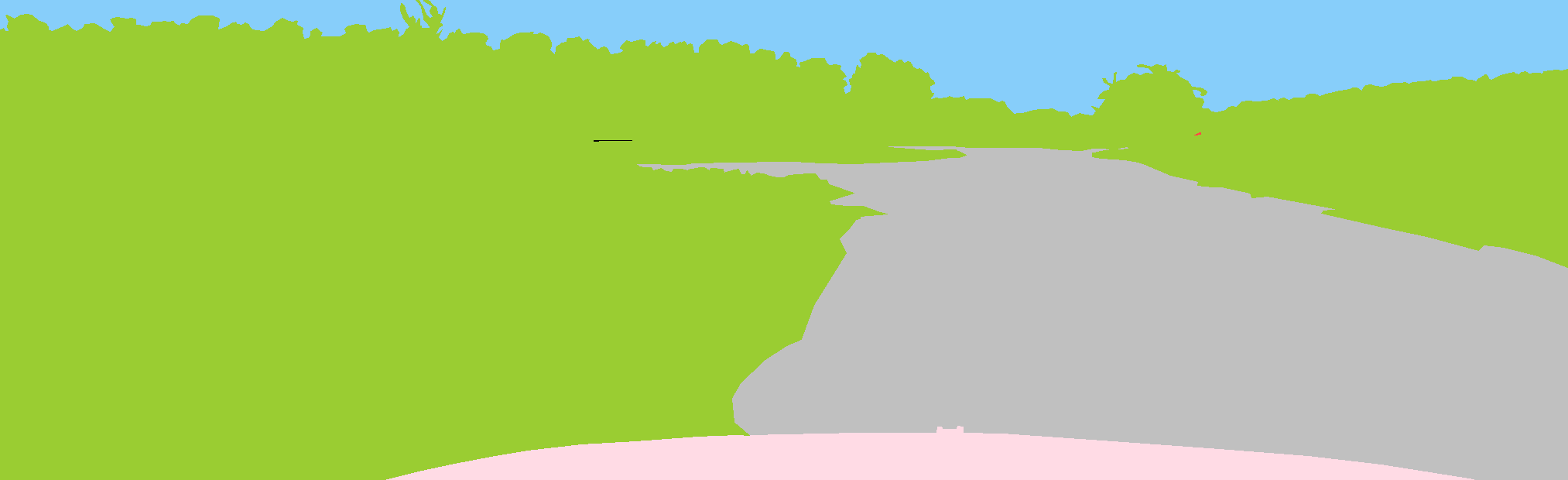} }}
	\hspace*{-0.45em}
	\subfloat{{\includegraphics[width=0.2477\textwidth]{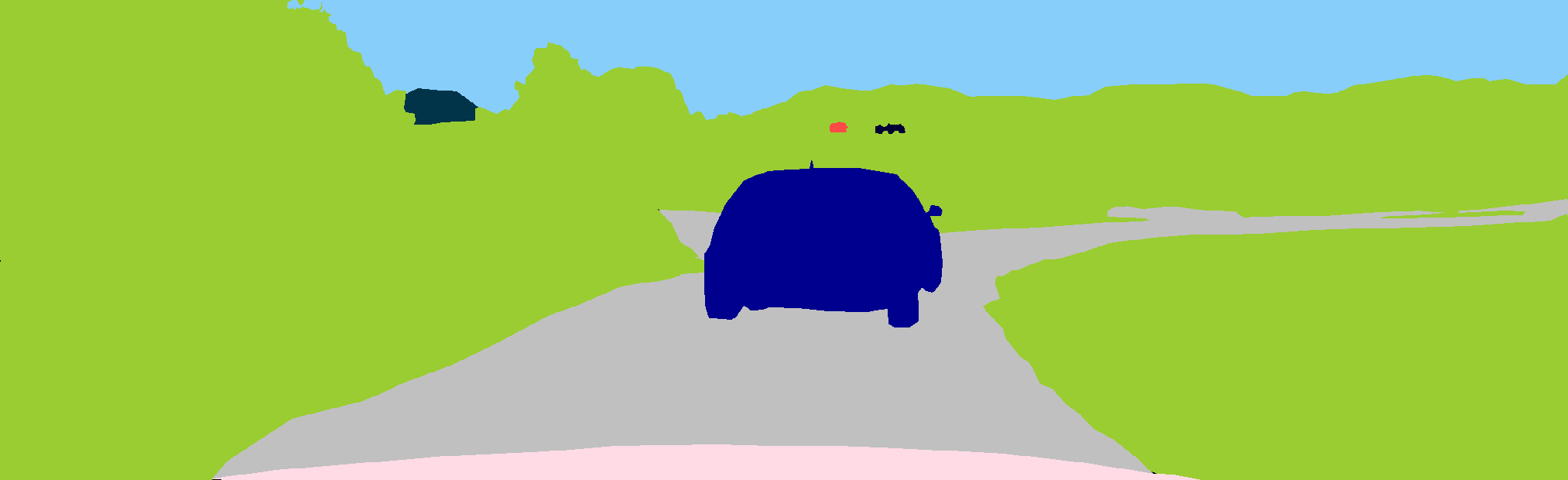} }}
	\hspace*{-0.45em}
	\subfloat{{\includegraphics[width=0.2477\textwidth]{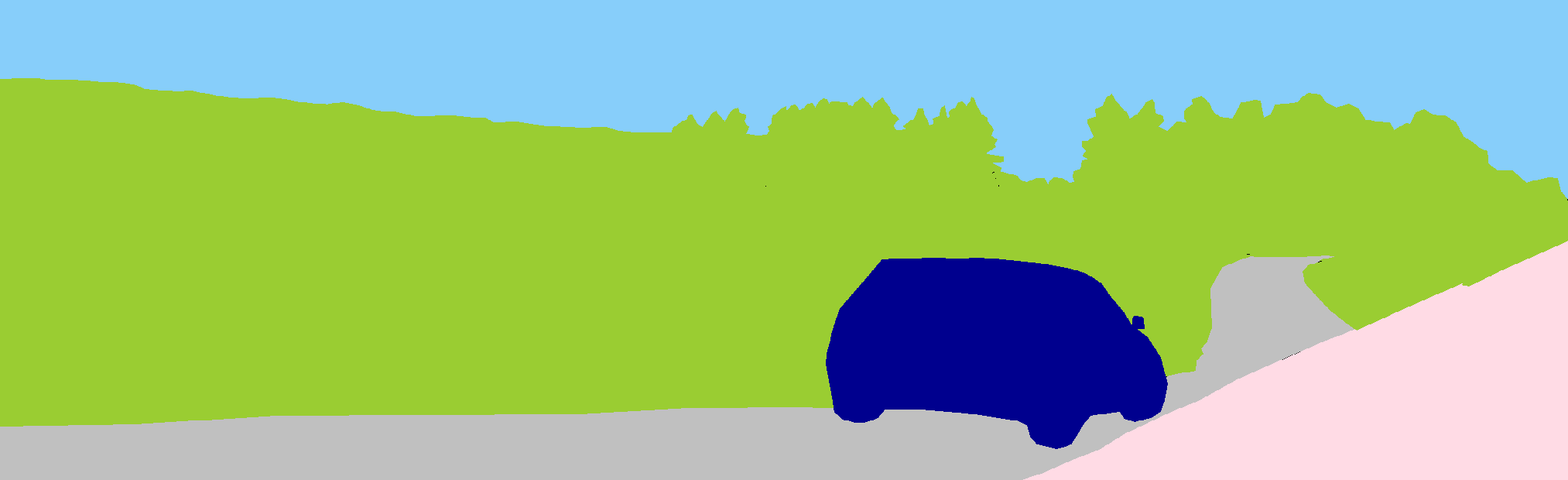} }}
	\hspace*{-0.45em}
	\subfloat{{\includegraphics[width=0.2477\textwidth]{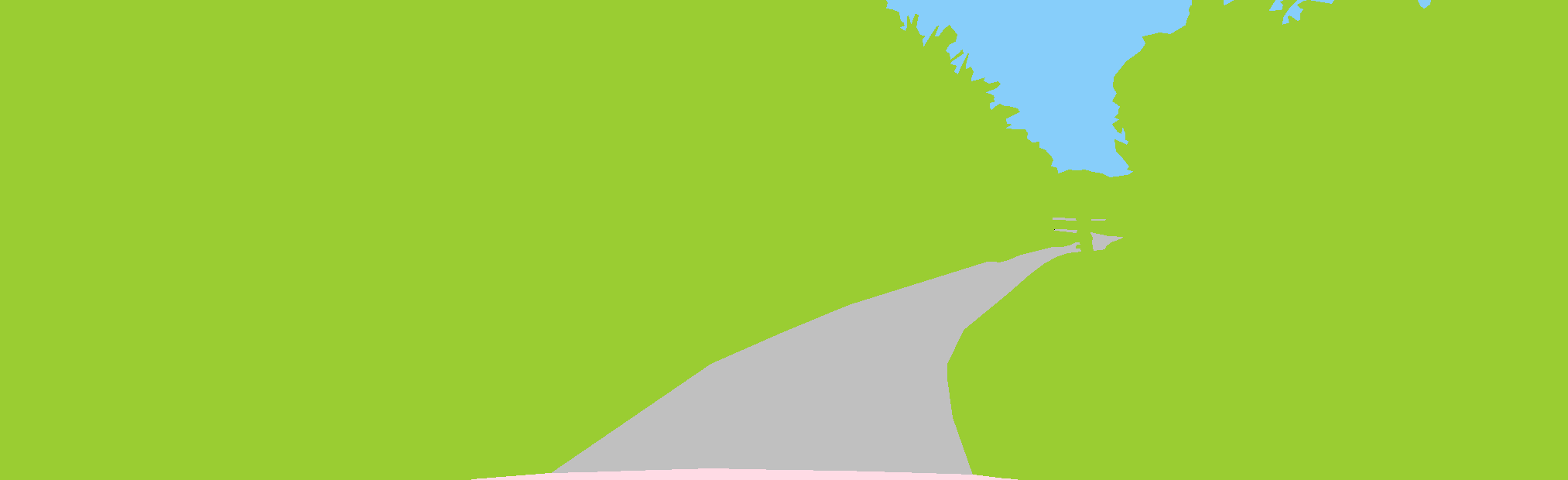} }}
	\\\vspace*{-0.8em}
	\subfloat{{\includegraphics[width=0.2477\textwidth]{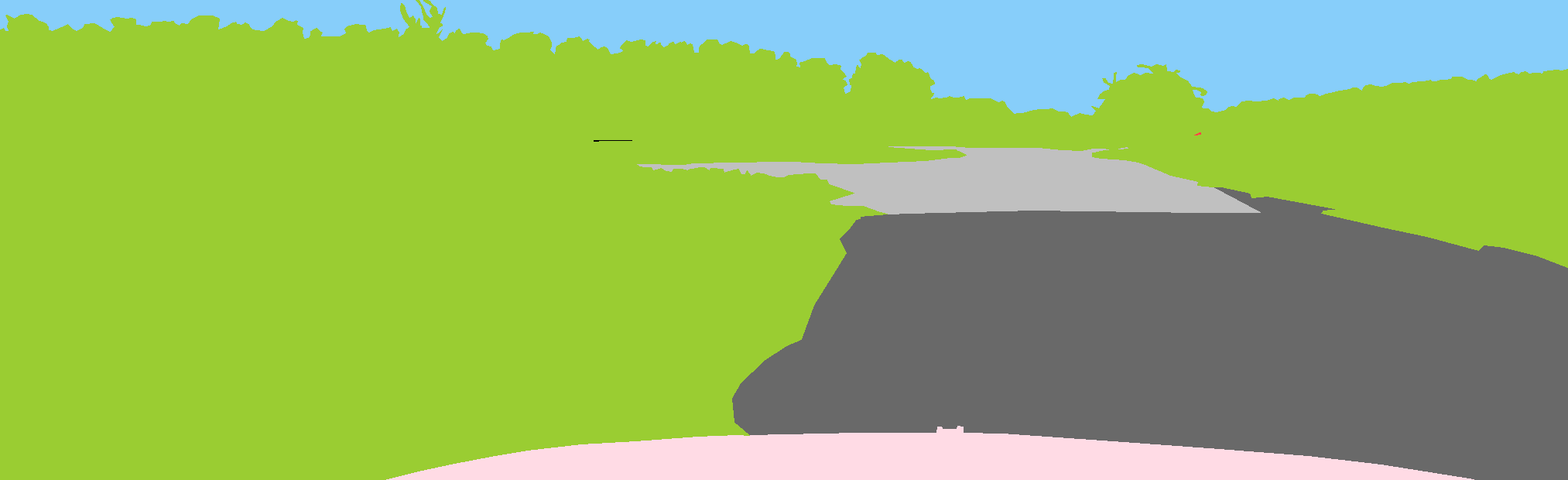} }}
	\hspace*{-0.45em}
	\subfloat{{\includegraphics[width=0.2477\textwidth]{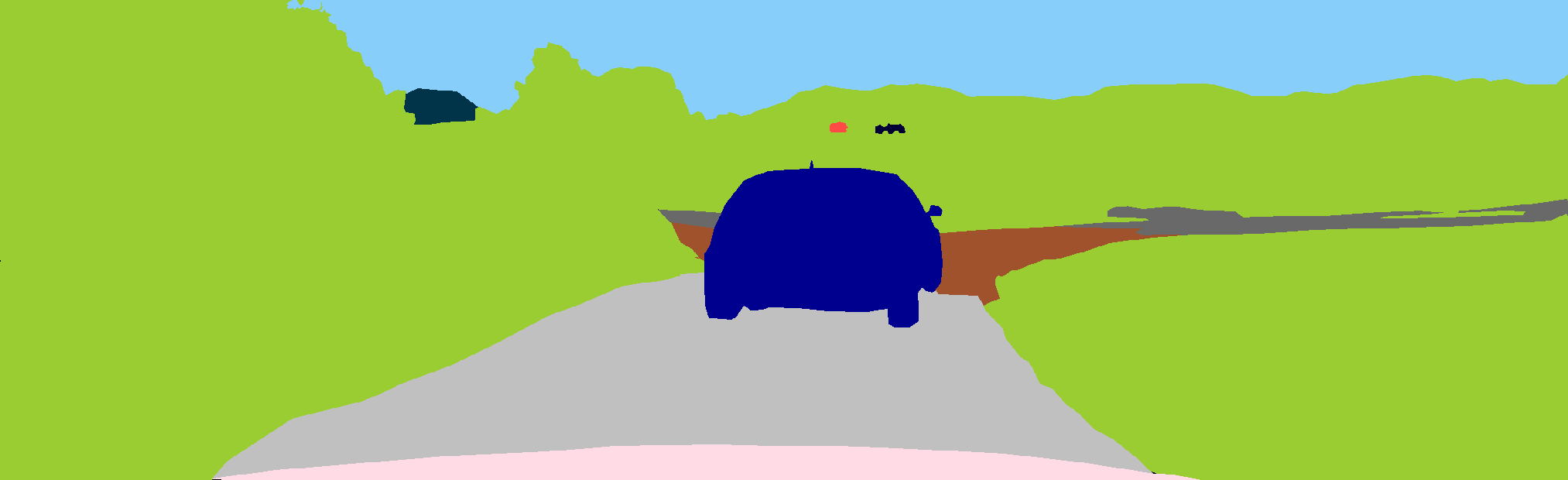} }}
	\hspace*{-0.45em}
	\subfloat{{\includegraphics[width=0.2477\textwidth]{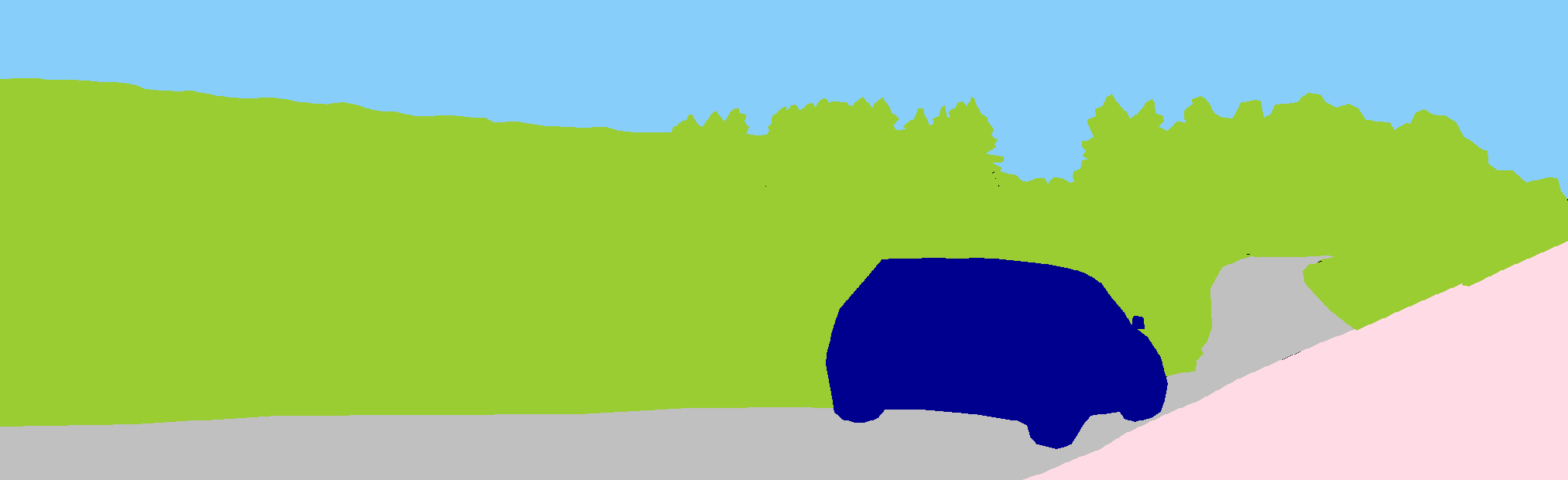} }}
	\hspace*{-0.45em}
	\subfloat{{\includegraphics[width=0.2477\textwidth]{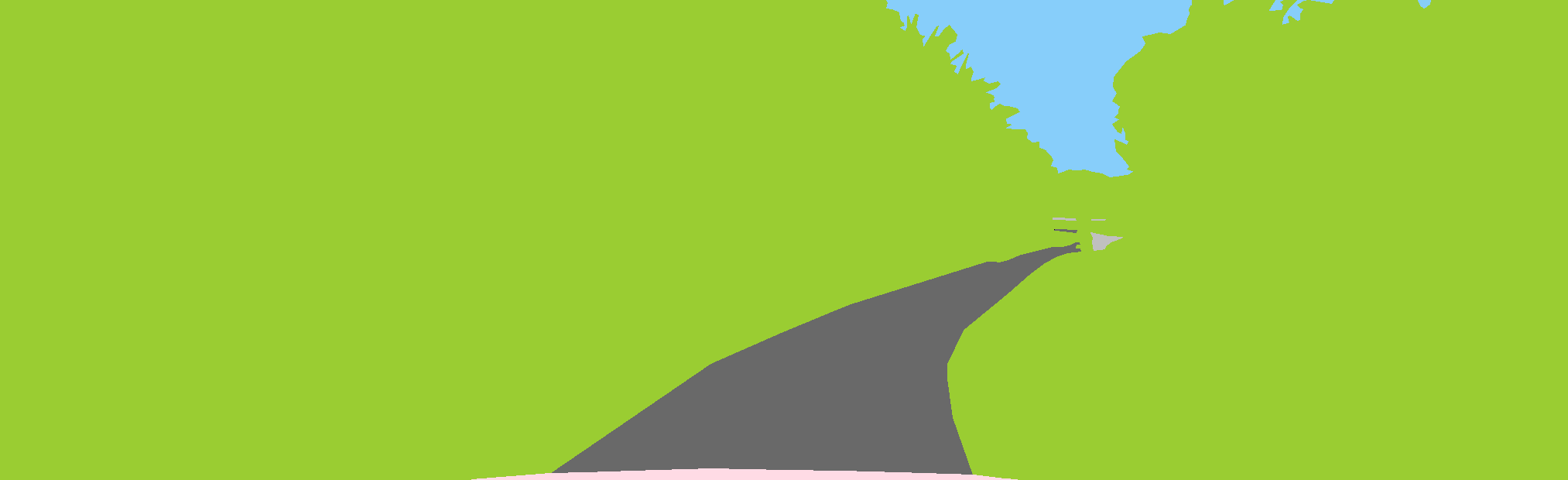} }}
	\\\vspace*{-0.8em}
	\subfloat{{\includegraphics[width=0.2477\textwidth]{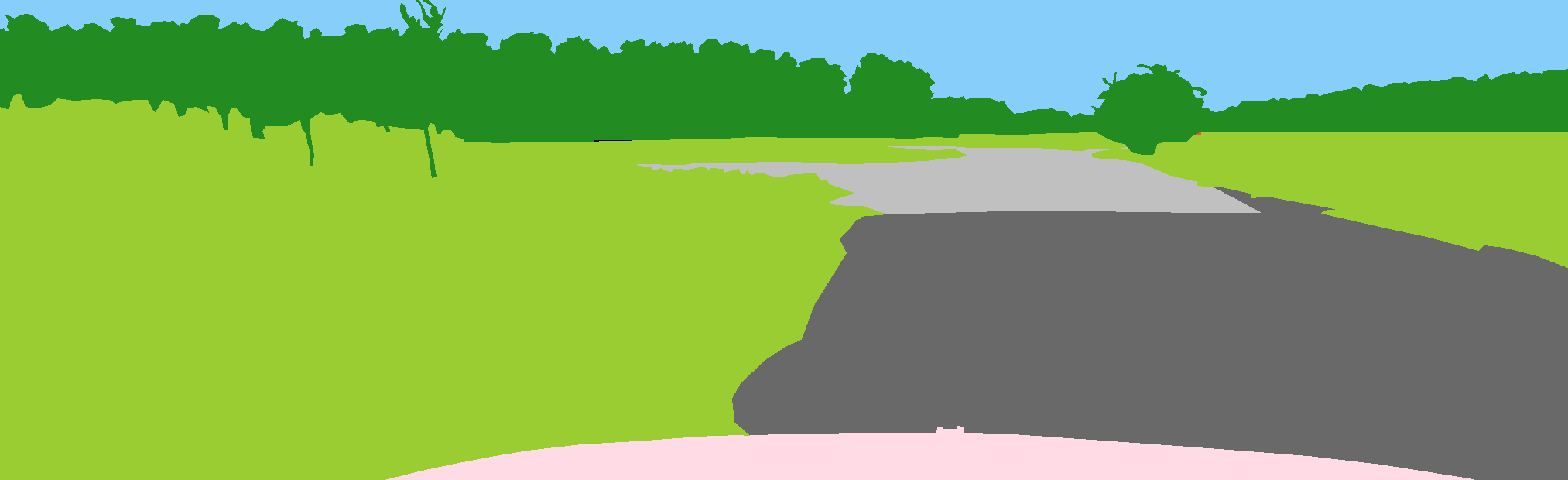} }}
	\hspace*{-0.45em}
	\subfloat{{\includegraphics[width=0.2477\textwidth]{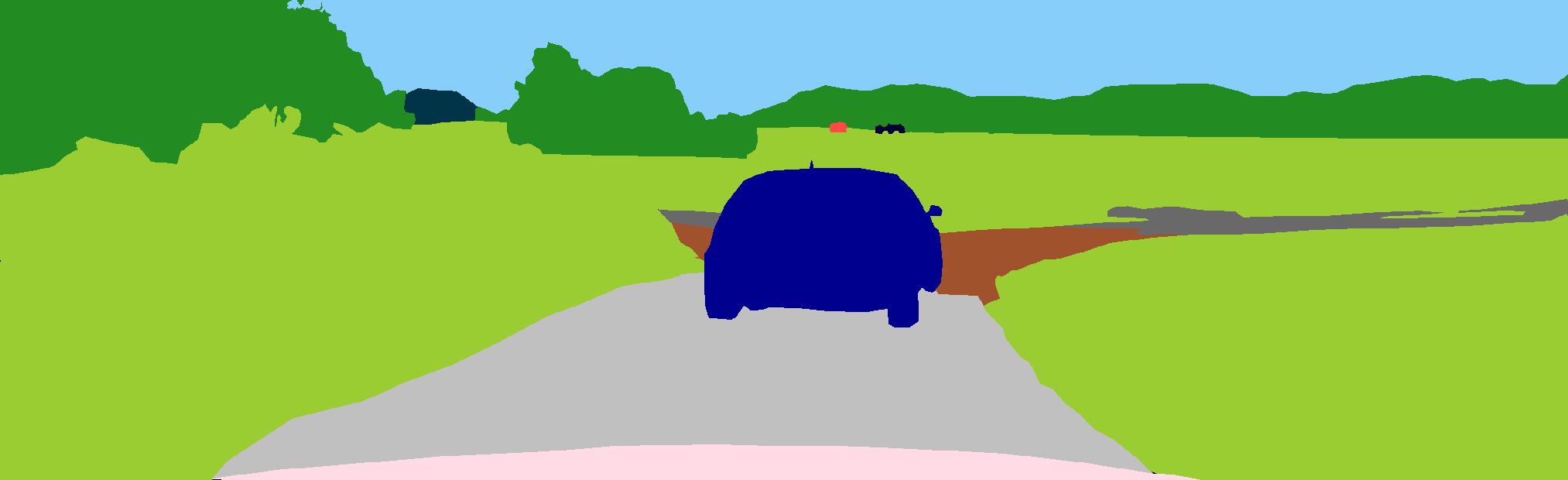} }}
	\hspace*{-0.45em}
	\subfloat{{\includegraphics[width=0.2477\textwidth]{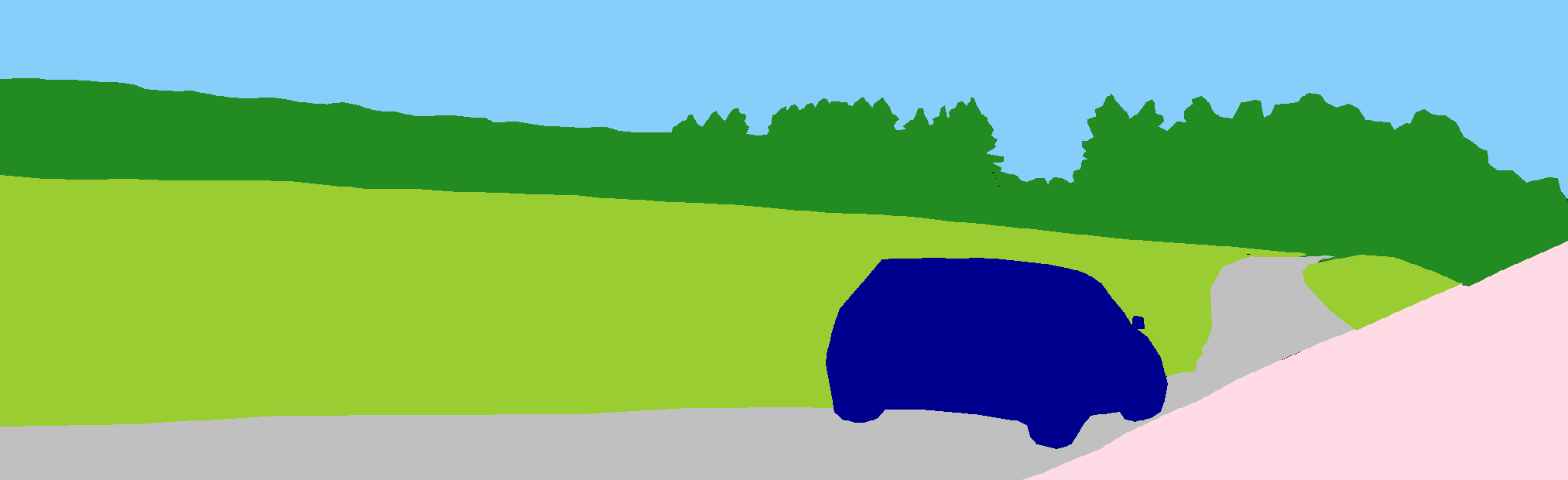} }}
	\hspace*{-0.45em}
	\subfloat{{\includegraphics[width=0.2477\textwidth]{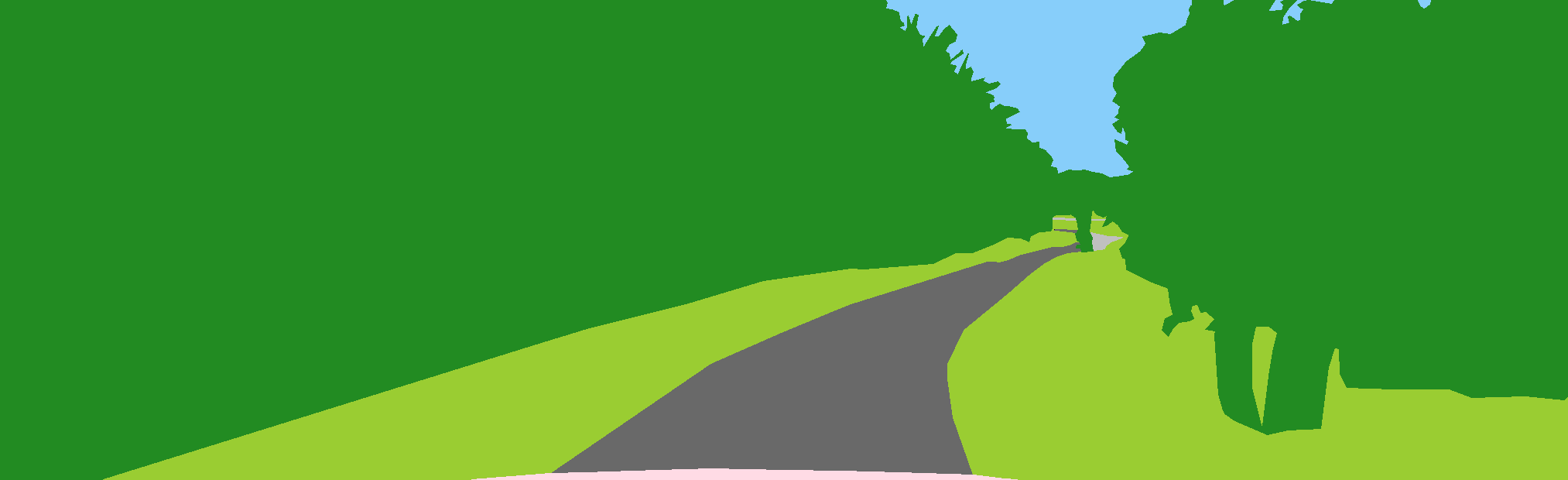} }}
	\\\vspace*{-0.8em}
	\subfloat{{\includegraphics[width=0.2477\textwidth]{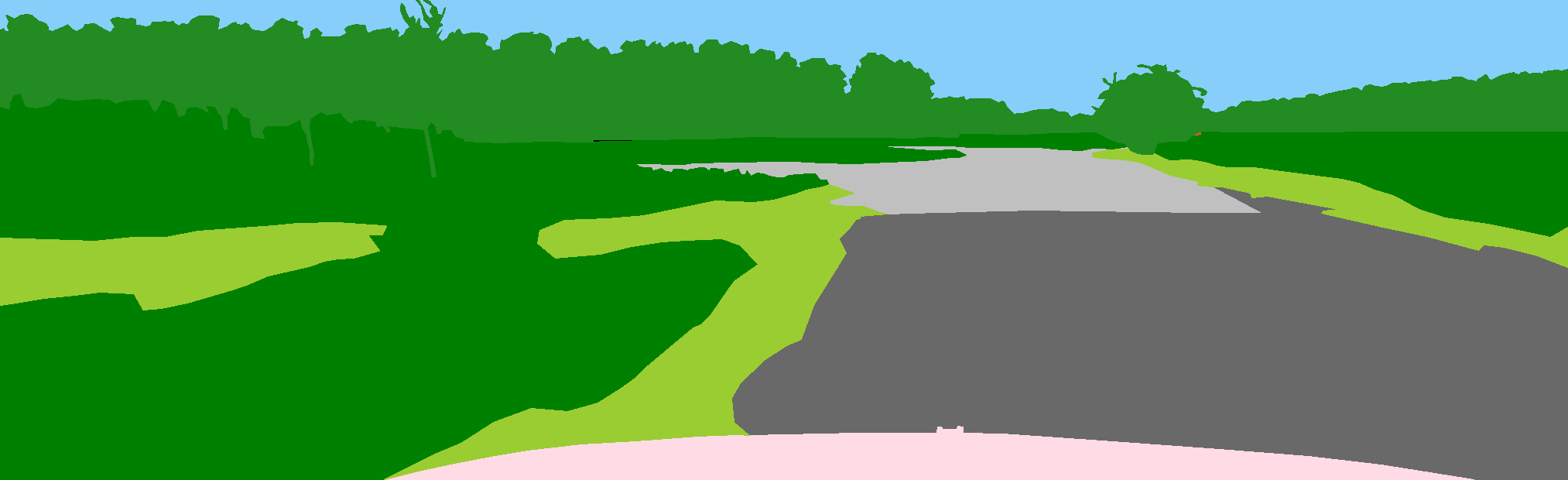} }}
	\hspace*{-0.45em}
	\subfloat{{\includegraphics[width=0.2477\textwidth]{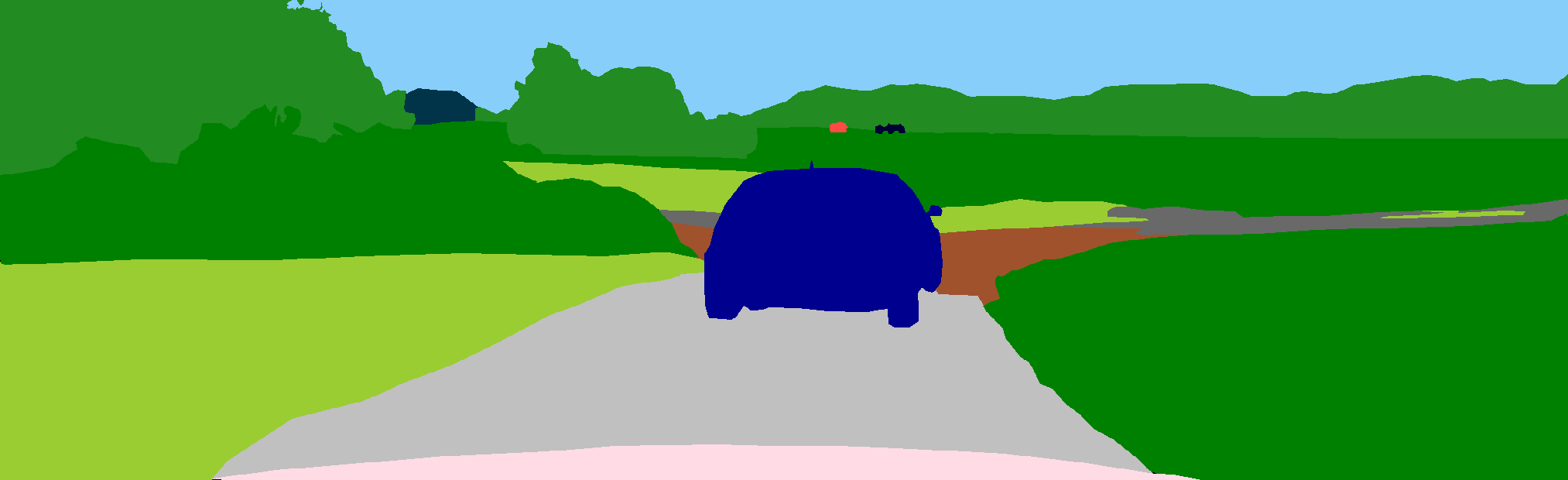} }}
	\hspace*{-0.45em}
	\subfloat{{\includegraphics[width=0.2477\textwidth]{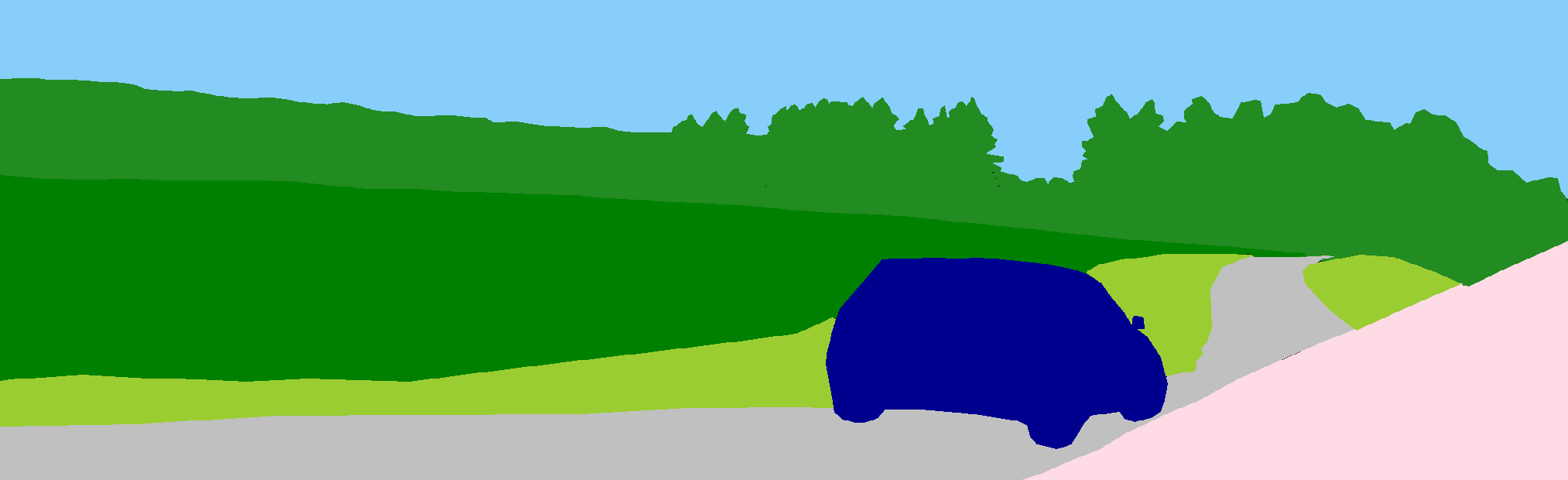} }}
	\hspace*{-0.45em}
	\subfloat{{\includegraphics[width=0.2477\textwidth]{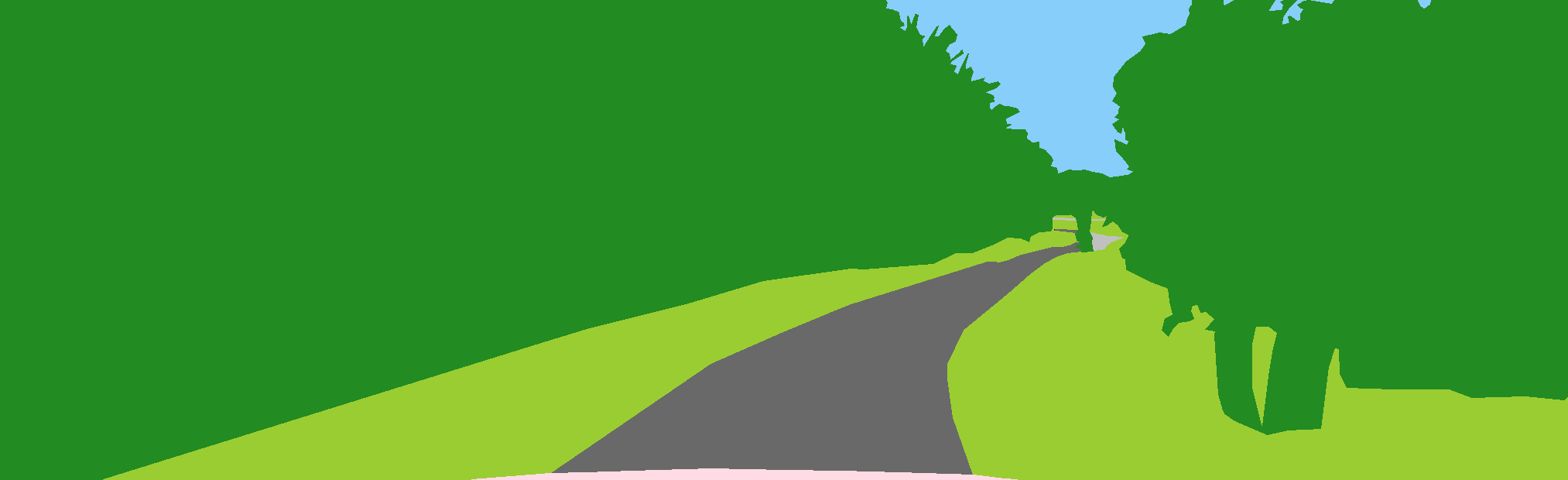} }}
	\\\vspace*{-0.8em}
	\subfloat{{\includegraphics[width=0.2477\textwidth]{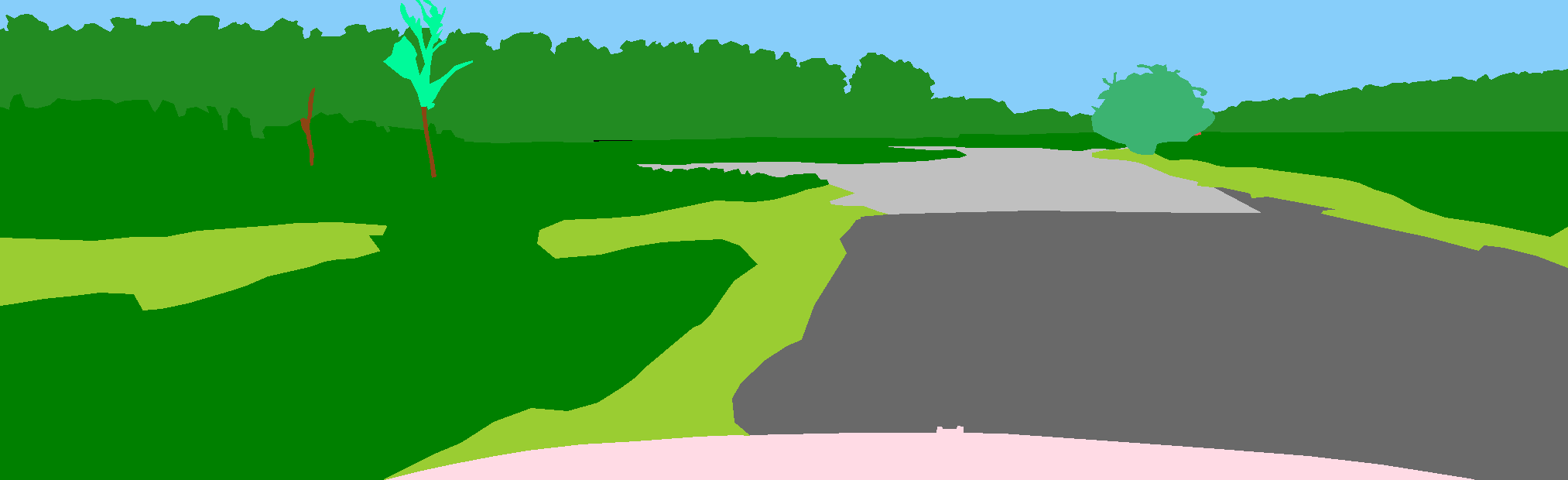} }}
	\hspace*{-0.45em}
	\subfloat{{\includegraphics[width=0.2477\textwidth]{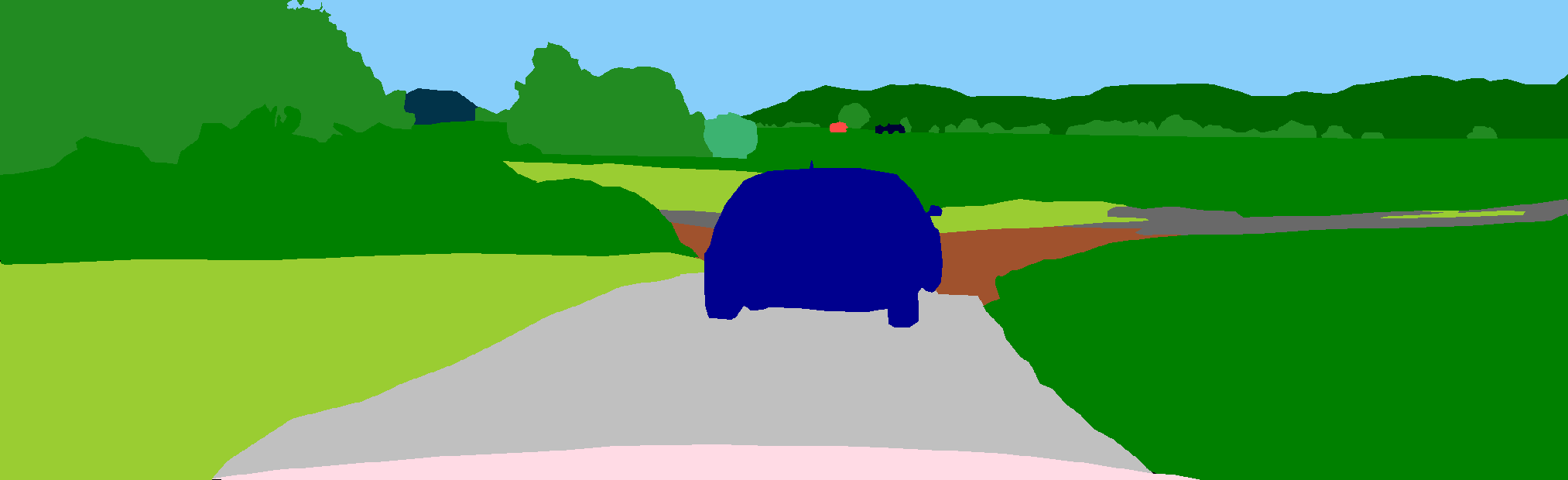} }}
	\hspace*{-0.45em}
	\subfloat{{\includegraphics[width=0.2477\textwidth]{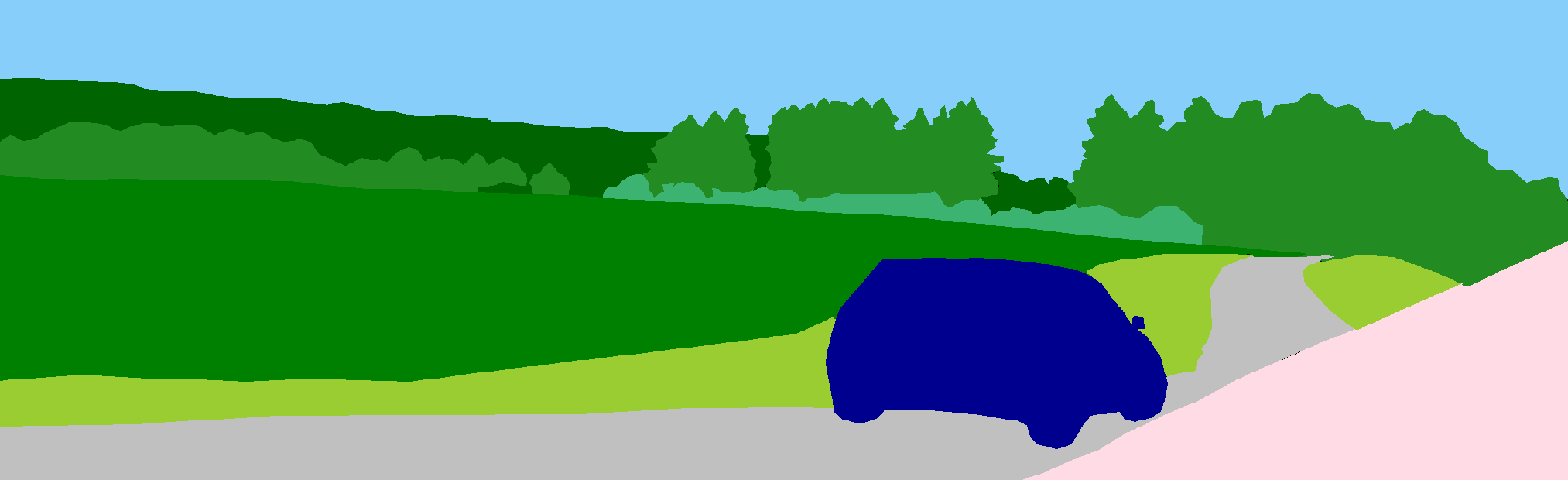} }}
	\hspace*{-0.45em}
	\subfloat{{\includegraphics[width=0.2477\textwidth]{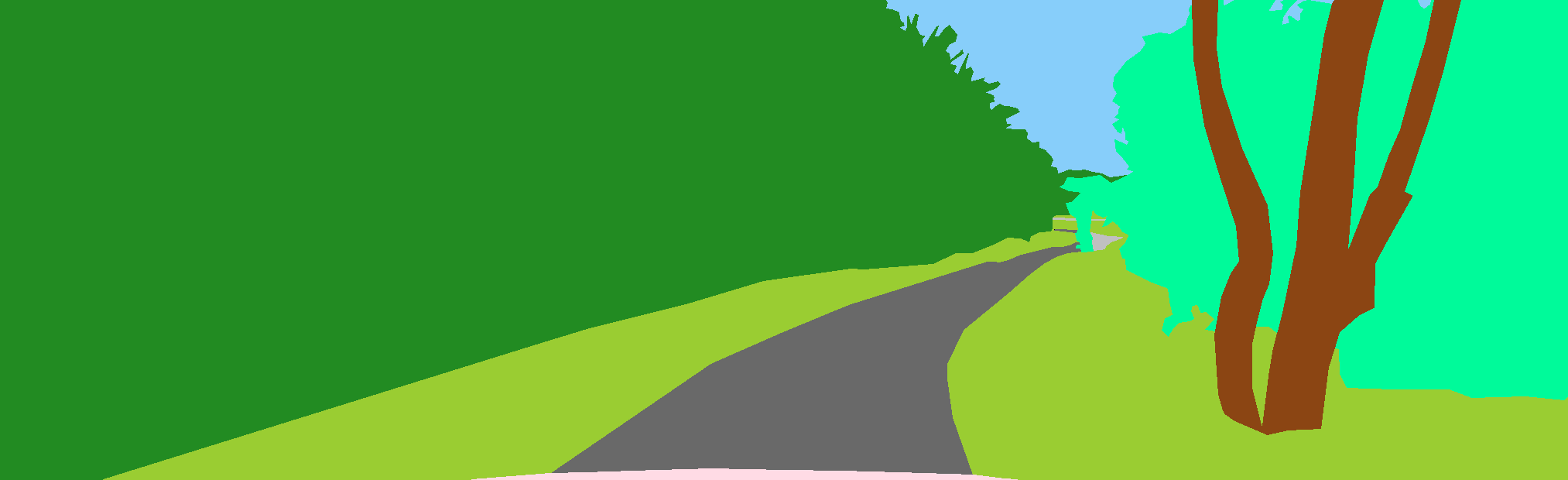} }}
	\caption{Examples of our proposed subdivision for terrain and vegetation categories. From top to bottom: image, annotation baseline with 14 classes following annotations with fine-grained level of granularity for 17, 18, 19 and 23 classes. 
	Our proposed subdivisions for terrain in gray (rows~$2 \rightarrow 3$); drivable vs. non-drivable vegetation (rows~$3 \rightarrow 4$), \textit{low} vs. \textit{high grass} (rows~$4 \rightarrow 5$), and fine-grained vegetation (rows~$5 \rightarrow 6$). The fine-grained vegetation in row 6 has a level of granularity that can distinguish between passable and impassable vegetation.
	The color coding for the semantic classes in the bottom row matches \Cref{fig:chp3:num-pixels}.}
	\label{fig:experiments:granularity}
\end{figure*}

\Cref{fig:experiments:granularity} shows examples of the proposed subdivision of our {\namedataset} dataset.
We annotated all images of our dataset pixel-wise and subdivided vegetation and terrain categories according to \Cref{fig:chp3:num-pixels}.
The baseline for this experiment are the condensed {\numcondensedclasses} classes from the previous experiment in \Cref{cpt5:section-comparison}.
First, the category terrain is split into \textit{asphalt}, \textit{gravel}, \textit{soil} and \textit{sand} (second row, \Cref{tab:ablation-granularity}, 17 classes).
Second, the category vegetation is subdivided into drivable (\textit{low} and \textit{high grass}) and non-drivable (\textit{bush}, \textit{forest}, \textit{tree crown}, \textit{tree trunk} and \textit{miscellaneous vegetation}) vegetation resulting in 18 classes.
Third, we further subdivide the drivable vegetation subgroup into \textit{low} and \textit{high grass} (19 classes).
\begin{table}[t]
	\centering
	\caption{Results when training Fast-SCNN for different levels of granularity. Values in brackets describe changes in performance with respect to the baseline with 14 classes. The best results on the validation set are in bold.}
	\label{tab:ablation-granularity}
	\begin{tabular}{@{}c|l|l|l@{}}
		\# C 
		& Level of granularity
		& Val mIoU (\%)
		& Val mBJ (\%) \\
		\hline
		\rule{0pt}{1.0\normalbaselineskip}
		14
		& Baseline (terrain, vegetation)
		& 53.00
		& 52.79 \\
		\hline\rule{0pt}{1.0\normalbaselineskip}
		\multirow{2}{*}{17}
		& Terrain
		&\multirow{2}{*}{53.54 (+0.54)}
		& \multirow{2}{*}{\textbf{54.29 (+1.50)}}\\
		& $\rightarrow$ \textit{Asphalt}, \textit{gravel}, \textit{soil}, \textit{sand}
		& &\\
		\hline\rule{0pt}{1.0\normalbaselineskip}
		\multirow{2}{*}{18}
		& Vegetation
		&\multirow{2}{*}{\textbf{54.77 (+1.77)}}
		& \multirow{2}{*}{53.56 (+0.77)}\\
		& $\rightarrow$ \textit{Drivable}, \textit{non-drivable} & &\\ 
		\hline\rule{0pt}{1.0\normalbaselineskip}
		\multirow{2}{*}{19}
		& Drivable vegetation
		&\multirow{2}{*}{52.30 (-0.70)}
		& \multirow{2}{*}{42.88 (-9.91)}\\
		& $\rightarrow$ \textit{Low grass}, \textit{high grass} & &\\
		\hline\rule{0pt}{1.0\normalbaselineskip}
		\multirow{3}{*}{23}
		& Non-drivable vegetation
		&\multirow{3}{*}{49.70 (-3.30)}
		& \multirow{3}{*}{50.12 (-2.67)}\\
		& $\rightarrow$ \textit{Bush}, \textit{forest}, \textit{misc. vegetation},  & &\\
		& $\quad$ \textit{tree crown}, \textit{tree trunk}  & &\\
	\end{tabular}
\end{table}
The subdivision of the drivable vegetation into \textit{low grass} and \textit{high grass} allows for a more detailed understanding of optimal paths along grassy fields. 
We, therefore, set a \SI{20}{\centi\meter} height threshold to distinguish between low and high grass.
Finally, the non-drivable vegetation subgroup is divided into the five fine-grained vegetation annotations \textit{bush}, \textit{forest}, \textit{tree crown}, \textit{tree trunk}, and \textit{miscellaneous vegetation},  arriving at 23 total classes.
The \textit{tree crown} and \textit{tree trunk} are annotated for isolated trees.
\Cref{tab:ablation-granularity} reports the performance for all levels of granularity (14 -- 23 classes) in terms of validation mIoU and mBJ, respectively, and also the improvements with respect to the baseline.

The third level of granularity (18 classes) is used from now on because the last two subdivisions (19 and 23 classes) do not improve the overall semantic segmentation performance.
This decline in segmentation performance with increasing class granularity could be due to very subtle visual differences between the fine-grained vegetation classes and the comparatively small training set size.
The annotation process itself might also be problematic because we use polygons to annotate structures in a pixel-wise manner, and this might be insufficiently precise for such intricate vegetation classes.
\subsection{Image Data Augmentation}
Image data augmentation has been applied in deep learning to reduce the dependence on large datasets and to reduce overfitting.
We want to explore label-preserving data augmentation techniques to enlarge the data-space for our {\namedataset} dataset.
Geometric transforms and color transformations were successfully tested on object classification as well as on semantic segmentation tasks \cite{kame:krizhevsky2012imagenet, kame:long2015fully, kame:casado2019clodsa}.
We employ random left-right flipping and random scaling on multiple scales for every training image of our dataset. 
\subsection{Ablation Study}
Using Fast-SCNN, we apply the following techniques to improve the classification performance on the test set of {\namedataset}. 
The test mIoU and mBJ scores for the ablation experiments are shown in \Cref{tab:ablation-fastscnn}.
\begin{itemize}
	\item \textbf{OHEM:}
	We utilize the online hard example mining (OHEM) loss according to \cite{kame:shrivastava2016training}. The OHEM loss function uses exclusively hard pixels to calculate the gradient for backpropagation. Prediction probabilities smaller than the threshold $\alpha = 0.7$ for correct classes are considered as hard examples.
	Only $\kappa = \frac{\text{bs} \cdot \text{h} \cdot \text{w}}{16}$ pixels are retained for each mini-batch, and those pixels are hard pixels.
	The performance for Fast-SCNN on our dataset improves by $+1.51$ mIoU on the test set.
	\item \textbf{Granularity:}
	The utilization of fine-grained class labels for terrain as well as for drivable and non-drivable vegetation improves the mIoU to $54.09$ w.r.t. the baseline. The mIoU drops compared to a coarser granularity level (w/ OHEM and w/o Gran.) on the test set. The mBJ increases with finer granularity to $57.23$.
	\item \textbf{Augmentations:}
	Applying left-right flipping and multi-scale augmentations (0.5x -- 2.0x) improves the performance to $65.26$ mIoU and $61.07$ mBJ on the test set.
	\item \textbf{Training set:}
	Further performance improvements on the test set are achieved by training Fast-SCNN on the combined training and validation set resulting in $67.94$ mIoU and $62.98$ mBJ on the test set.
\end{itemize}
\begin{figure*}[!t]%
	\setlength{\tempwidth}{.3\linewidth}
	\settoheight{\tempheight}{\includegraphics[width=\tempwidth]{example-image-a}}%
	\centering
	\textcolor{black}{\rule{\linewidth}{1pt}}
	{
		\vspace{-3mm}
		\columnname{Input Image}\hfil
		\columnname{Ground Truth}\hfil
		\columnname{Fast-SCNN Prediction}
	}
	\subfloat{{\includegraphics[width=0.3309\textwidth]{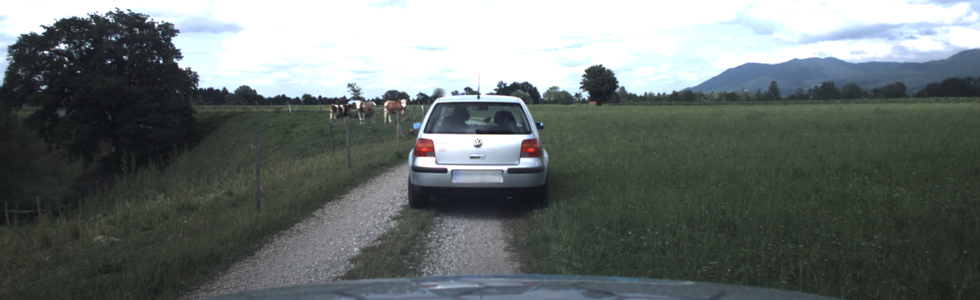} }}
	\hspace*{-0.45em}
	\subfloat{{\includegraphics[width=0.3309\textwidth]{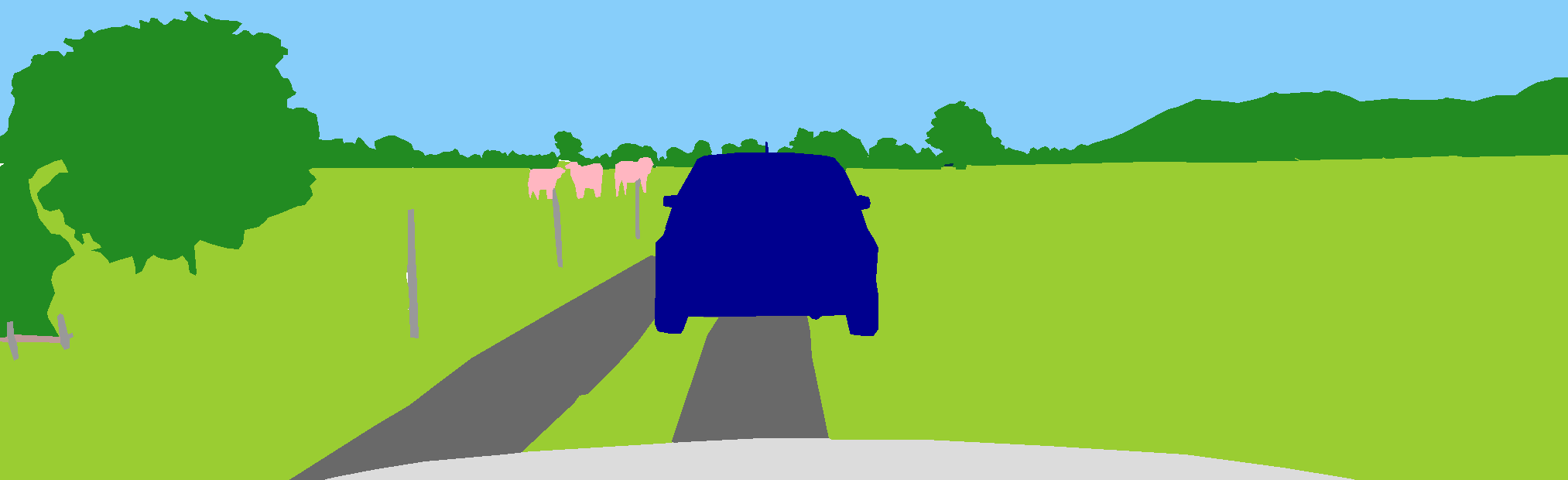} }}
	\hspace*{-0.45em}
	\subfloat{{\includegraphics[width=0.3309\textwidth]{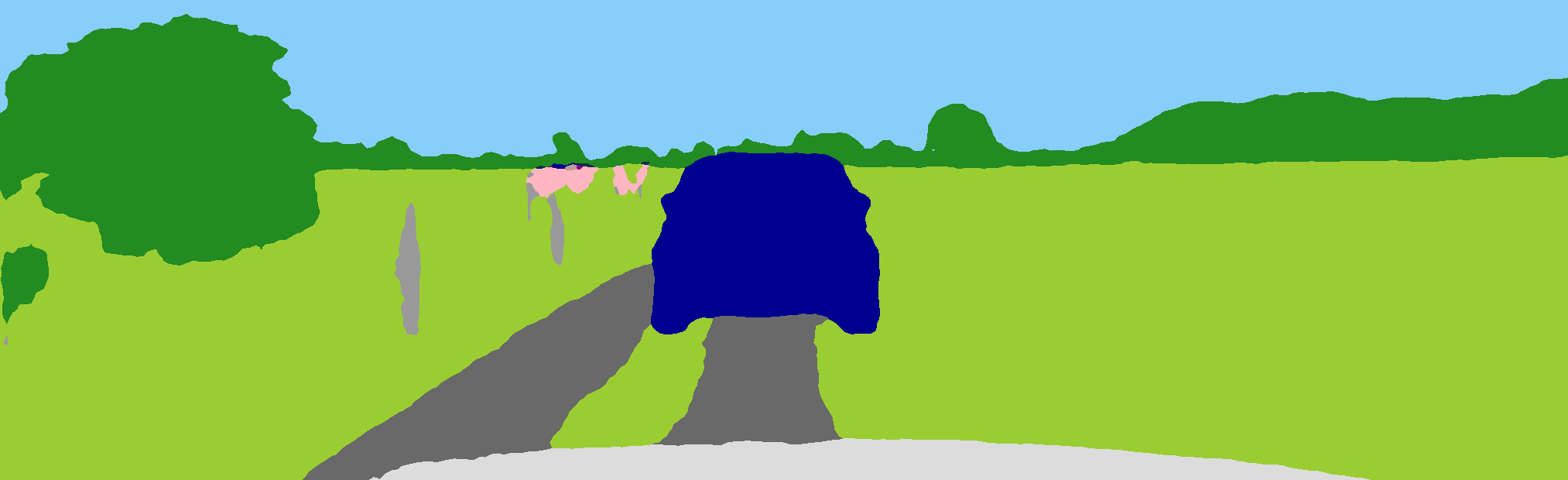} }}
	\\\vspace*{-0.8em}
	\subfloat{{\includegraphics[width=0.3309\textwidth]{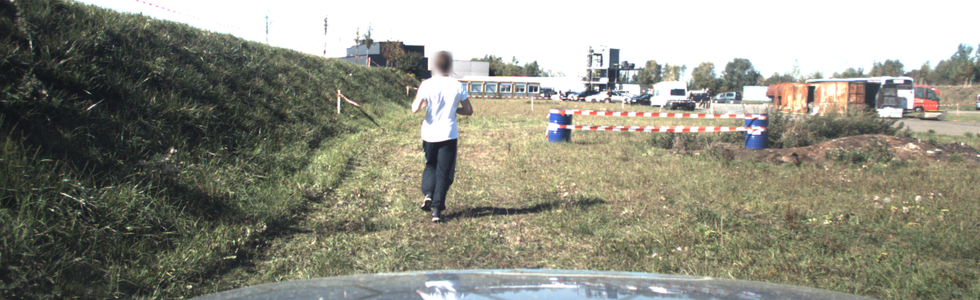} }}
	\hspace*{-0.45em}
	\subfloat{{\includegraphics[width=0.3309\textwidth]{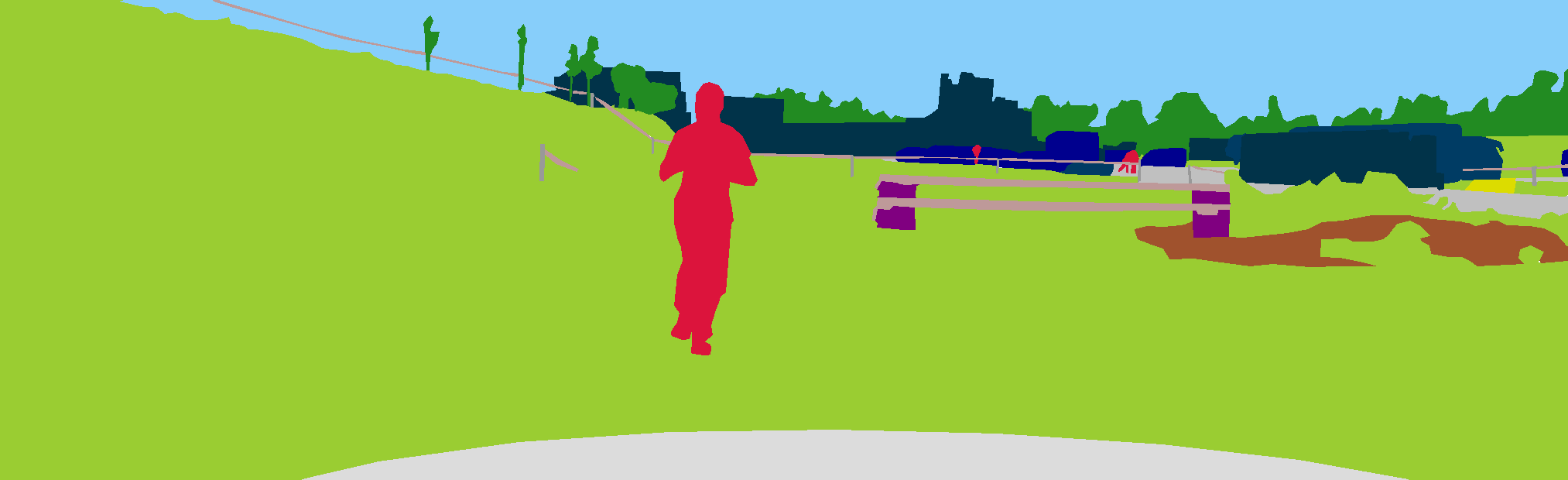} }}
	\hspace*{-0.45em}
	\subfloat{{\includegraphics[width=0.3309\textwidth]{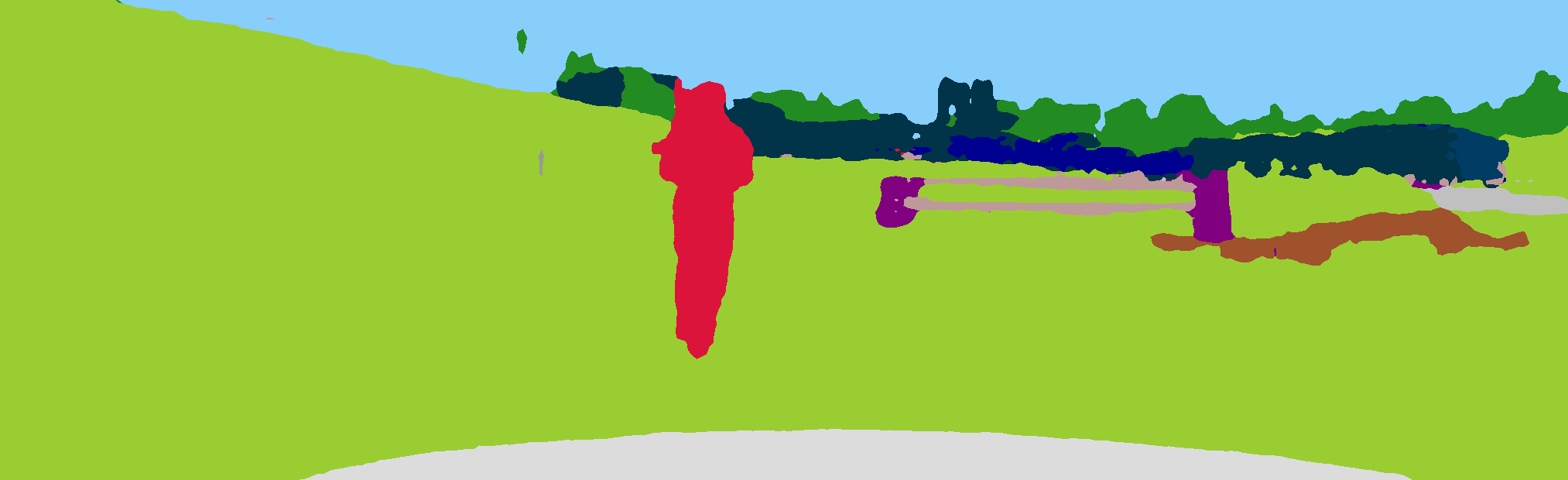} }}
	\\\vspace*{-0.8em}
	\subfloat{{\includegraphics[width=0.3309\textwidth]{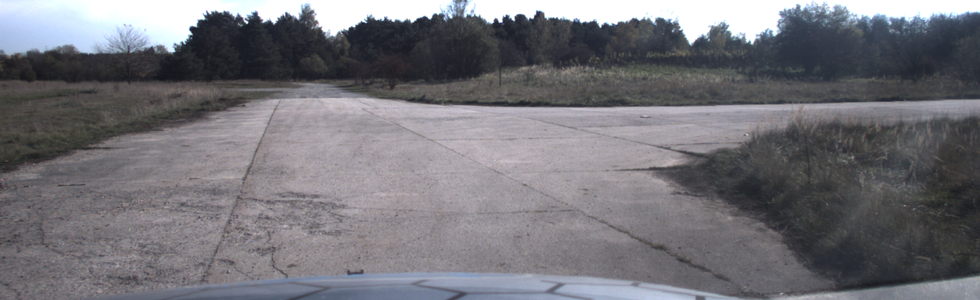} }}	
	\hspace*{-0.45em}
	\subfloat{{\includegraphics[width=0.3309\textwidth]{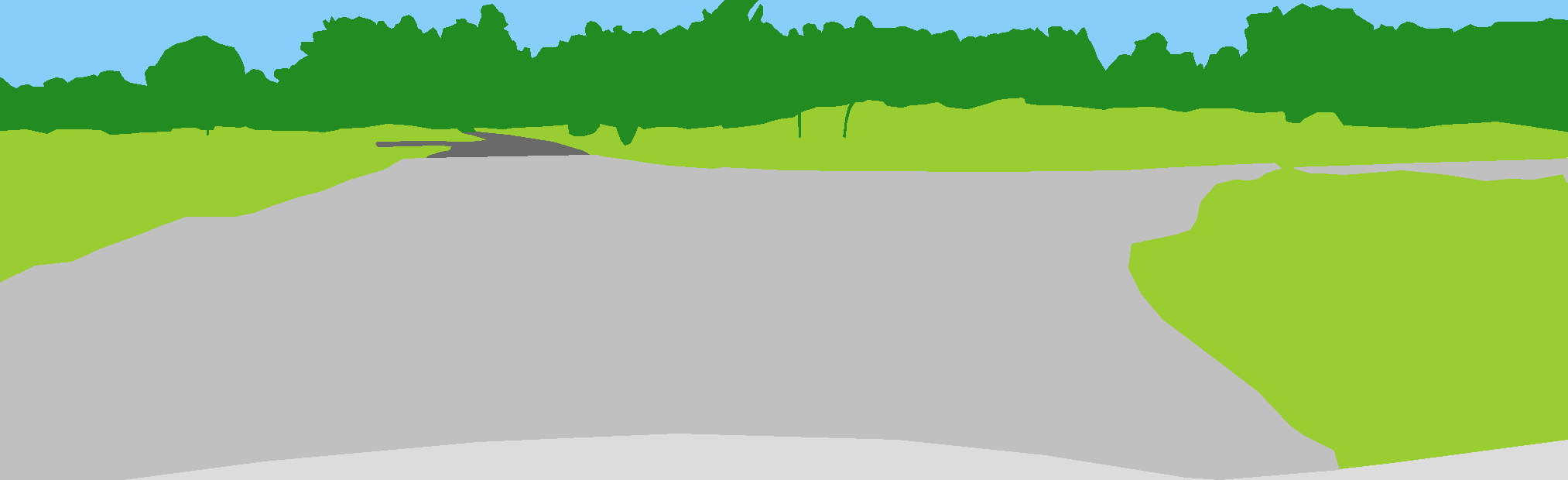} }}
	\hspace*{-0.45em}
	\subfloat{{\includegraphics[width=0.3309\textwidth]{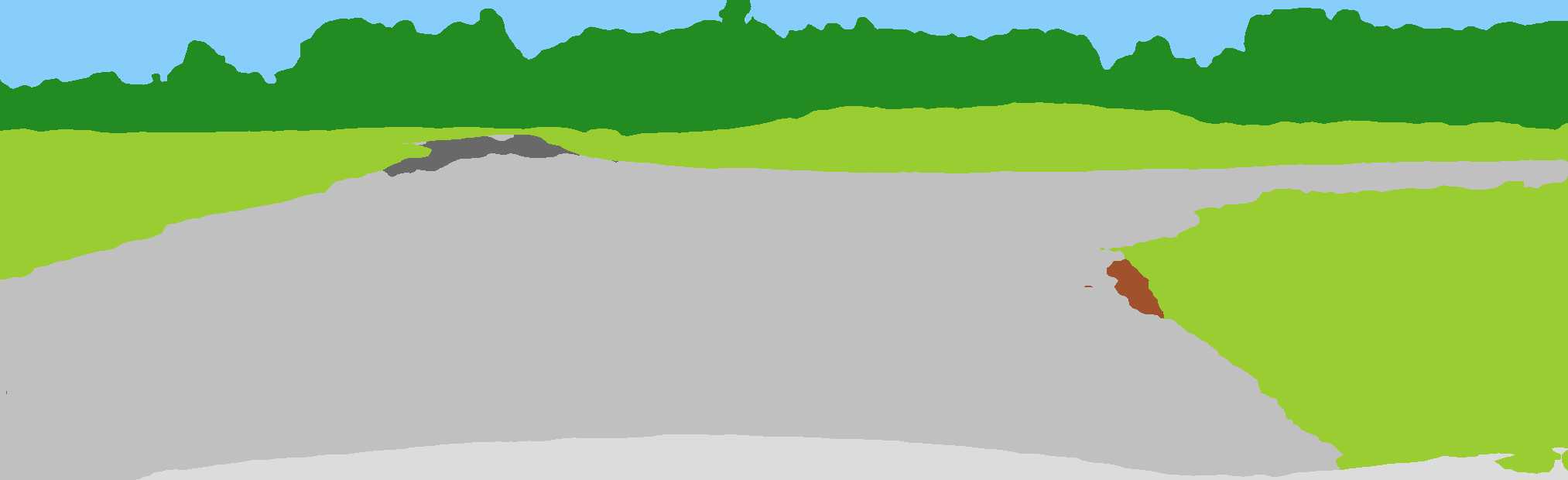} }}
	\\\vspace*{-0.8em}
	\subfloat{{\includegraphics[width=0.3309\textwidth]{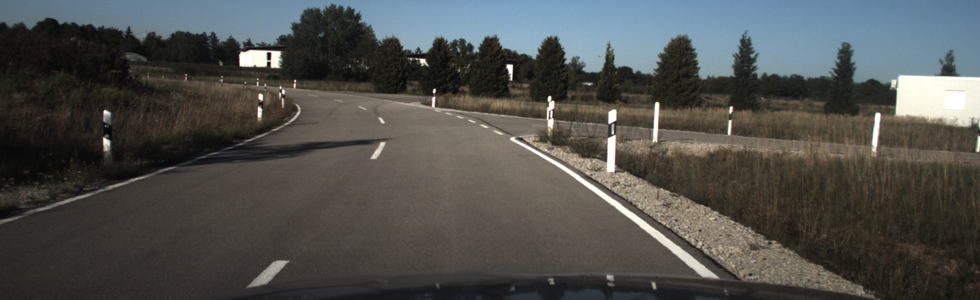} }}
	\hspace*{-0.45em}
	\subfloat{{\includegraphics[width=0.3309\textwidth]{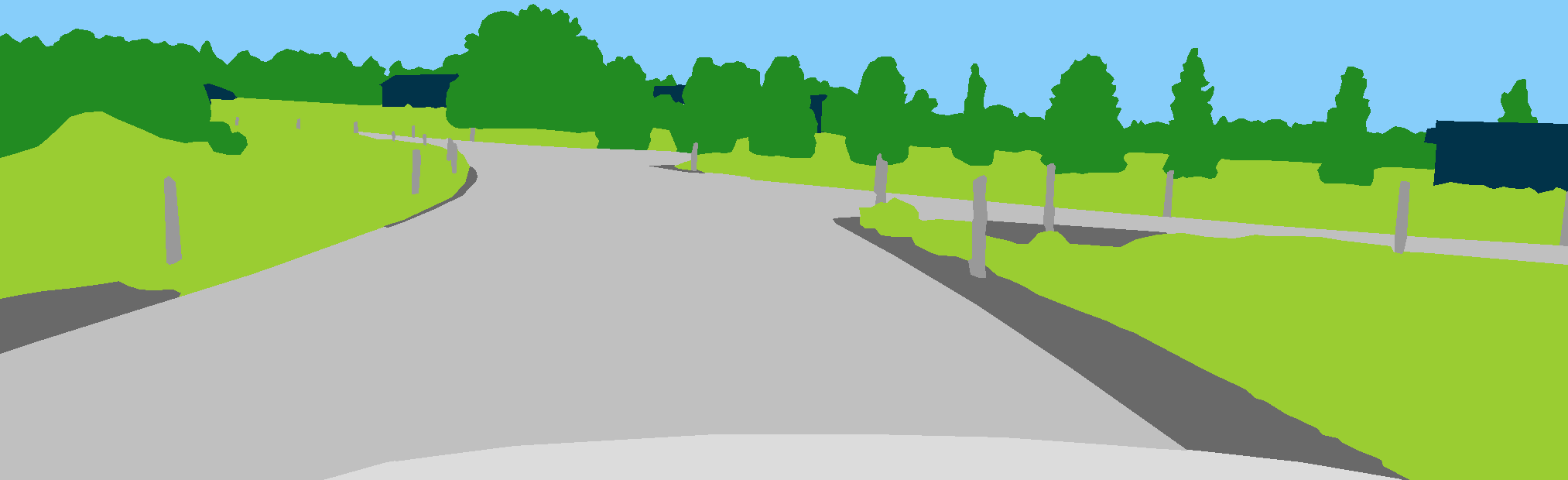} }}
	\hspace*{-0.45em}
	\subfloat{{\includegraphics[width=0.3309\textwidth]{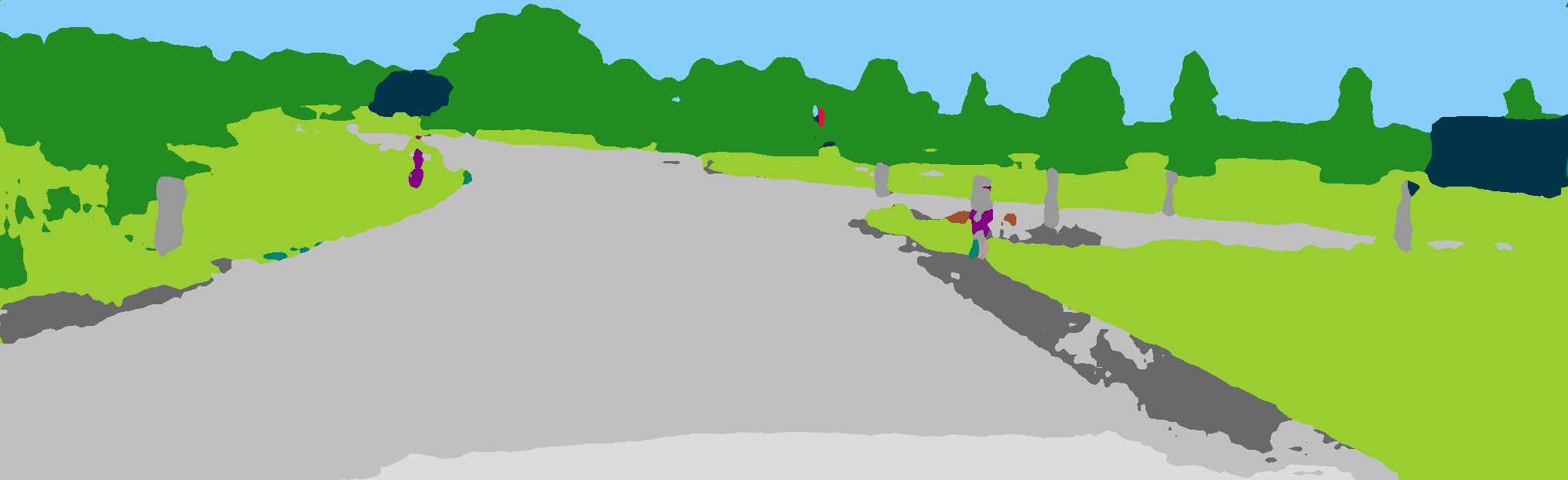} }}
	\\\vspace*{-0.8em}
	\subfloat{{\includegraphics[width=0.3309\textwidth]{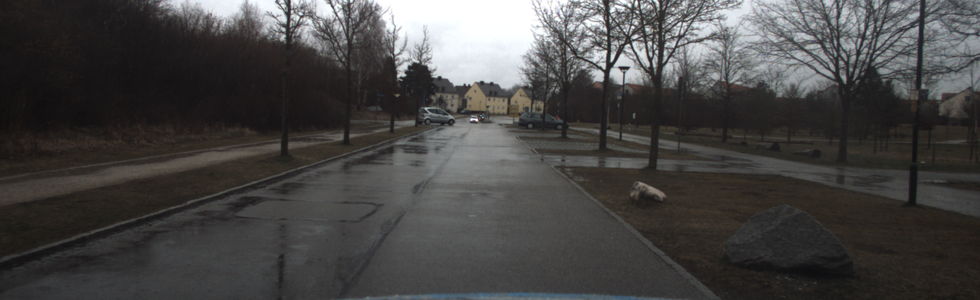} }}
	\hspace*{-0.45em}
	\subfloat{{\includegraphics[width=0.3309\textwidth]{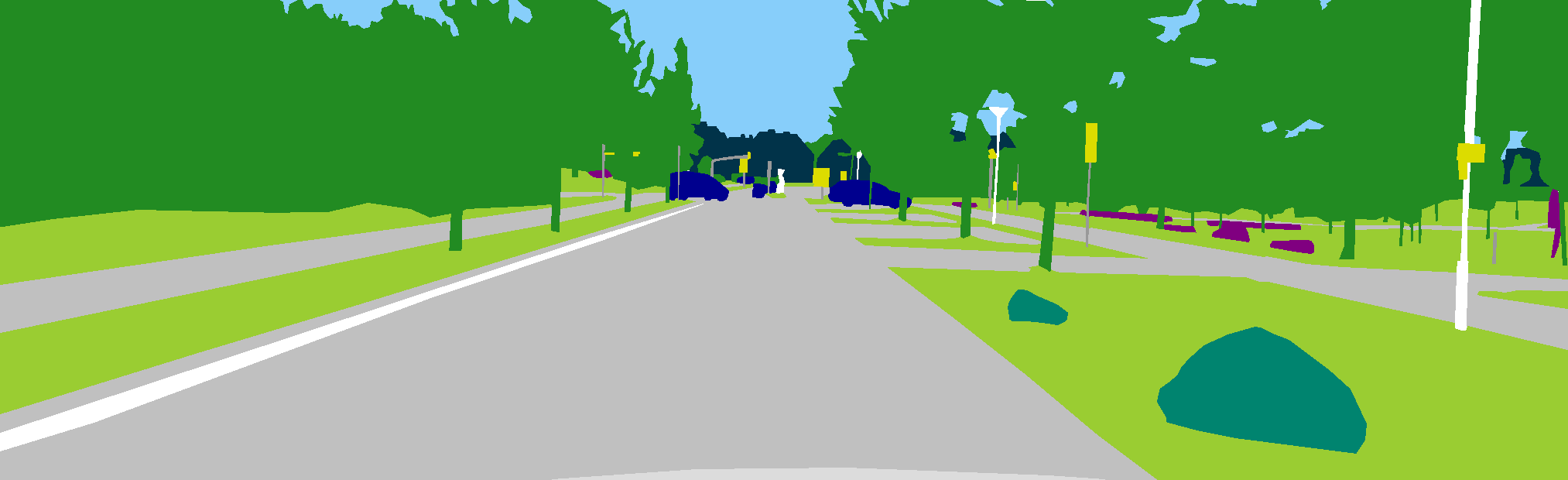} }}
	\hspace*{-0.45em}
	\subfloat{{\includegraphics[width=0.3309\textwidth]{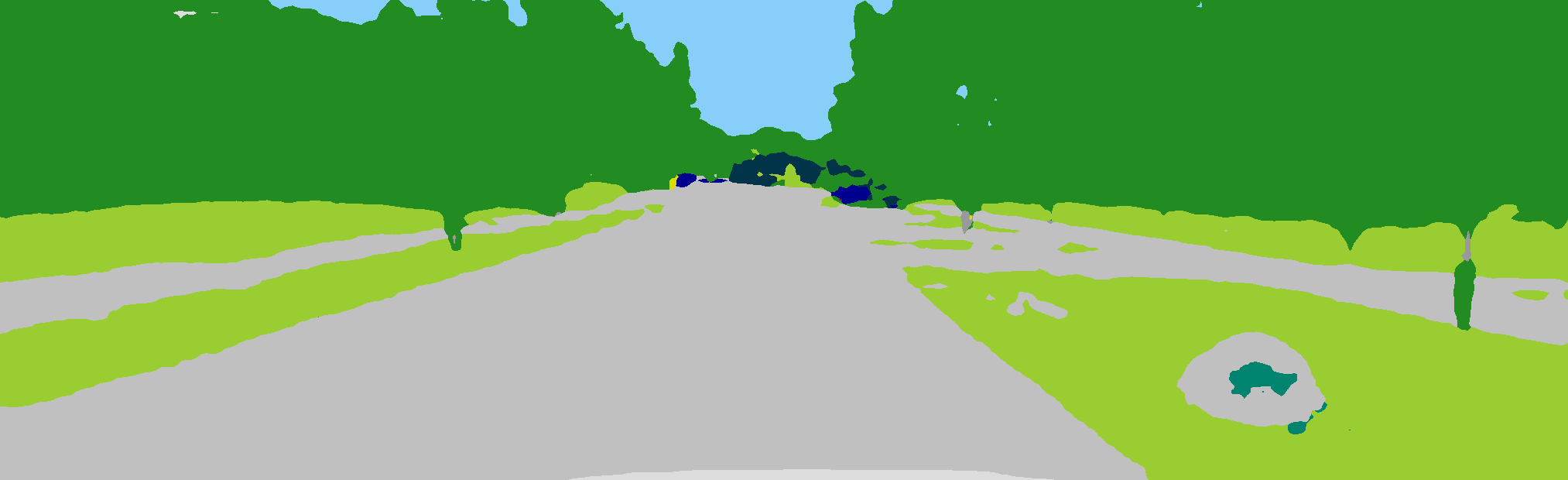} }}
	
	\textcolor{black}{\rule{\linewidth}{1pt}}
	\caption{Qualitative results of our ablation with the Fast-SCNN method. From left to right: image, ground truth mask and predicted semantic mask. The color coding for the semantic classes matches \Cref{fig:chp3:num-pixels} except we use two green tones for drivable and non-drivable vegetation.
	}
	\label{fig:experiments:quality}%
\end{figure*}

\begin{table}[t]
	\vspace{1.3mm}
	\centering
	\caption{We report the results of our ablation on the test set of the {\namedataset} dataset.}
	\label{tab:ablation-fastscnn}
	\begin{tabular}{@{}c|c|c|c|l|l@{}}
		\begin{tabular}[c]{@{}c@{}}OHEM\\\end{tabular}
		& \begin{tabular}[c]{@{}c@{}}Gran.\\\end{tabular}
		& \begin{tabular}[c]{@{}c@{}}Augm.\\\end{tabular}
		& \begin{tabular}[c]{@{}c@{}}Trainval\\\end{tabular}
		& \begin{tabular}[c]{@{}c@{}}Test mIoU ($\%$)\end{tabular}
		& \begin{tabular}[c]{@{}c@{}}Test mBJ ($\%$)\end{tabular}\\
		\hline
		\rule{0pt}{1.0\normalbaselineskip}
		\xmark &  \xmark &  \xmark  &\xmark  & 53.88 		 		& 51.06 \\
		\cmark &  \xmark &  \xmark  &\xmark  & 55.39 (+$\:\;$1.51) 	& 55.67 (+\hspace{1.45mm}4.61) \\ 
		\cmark &  \cmark &  \xmark  &\xmark  & 54.09 (+$\:\;$0.21) 	
		& 57.23 (+\hspace{1.45mm}6.17) \\
		\cmark &  \cmark &  \cmark  &\xmark  & 65.26 (+11.38) 		& 61.07 (+10.01) \\
		\cmark &  \cmark &  \cmark  &\cmark  & \textbf{67.94 (+14.06)}
		& \textbf{62.98 (+11.92)} \\
	\end{tabular}
\end{table}

\Cref{fig:experiments:quality} shows the qualitative results of our final Fast-SCNN model.
The classes \textit{pole}, \textit{tree trunks} and \textit{fence} that resemble thin obstacles are challenging and Fast-SCNN has problems detecting these classes reliably at greater distances (rows 1, 2, 4, and 5 in \Cref{fig:experiments:quality}).
There are also other underrepresented classes such as \textit{animal} (first row in \Cref{fig:experiments:quality}) and a rock (prediction in the fifth row in \Cref{fig:experiments:quality}).
The latter class might be partially confused with \textit{asphalt} because it shares a similar color and texture.
Overall, we observe that for the task of distinguishing vegetation from thing classes that have a different color and clear structures Fast-SCNN performs well. But for finer distinctions between vegetation and terrain types we require more semantically labeled high-resolution data.


\section{Discussion}
\label{sec:discussion}

Our evaluation of different network architectures emphasizes the trade-off between the output resolution, inference speed, and memory footprint that has to be considered when choosing a semantic segmentation architecture for mobile applications. 
We choose the Fast-SCNN method for effective image segmentation in unstructured driving scenarios.
The subdivision of terrain classes improves the overall segmentation performance to some extent. 
Promising results are demonstrated for similar scenes, but a larger dataset is required to accurately classify subclasses of vegetation during different seasons and weather conditions.
We believe the creation of driving datasets similar to {\namedataset} will further improve the visual perception in unstructured environments and eventually also for offroad navigation.

\bibliographystyle{IEEEtran}
\balance
\bibliography{bib/IEEEfull,bib/additional_abrv,bib/icpr_tas_segmentation,bib/et_al}

\begin{thebibliography}{10}
\providecommand{\url}[1]{#1}
\csname url@samestyle\endcsname
\providecommand{\newblock}{\relax}
\providecommand{\bibinfo}[2]{#2}
\providecommand{\BIBentrySTDinterwordspacing}{\spaceskip=0pt\relax}
\providecommand{\BIBentryALTinterwordstretchfactor}{4}
\providecommand{\BIBentryALTinterwordspacing}{\spaceskip=\fontdimen2\font plus
\BIBentryALTinterwordstretchfactor\fontdimen3\font minus
  \fontdimen4\font\relax}
\providecommand{\BIBforeignlanguage}[2]{{%
\expandafter\ifx\csname l@#1\endcsname\relax
\typeout{** WARNING: IEEEtran.bst: No hyphenation pattern has been}%
\typeout{** loaded for the language `#1'. Using the pattern for}%
\typeout{** the default language instead.}%
\else
\language=\csname l@#1\endcsname
\fi
#2}}
\providecommand{\BIBdecl}{\relax}
\BIBdecl

\bibitem{brostow2009semantic}
G.~J. Brostow, J.~Fauqueur, and R.~Cipolla, ``{Semantic Object Classes in
  Video: A High-Definition Ground Truth Database},'' \emph{Pattern Recognition
  Letters}, vol.~30, no.~2, pp. 88--97, 2009.

\bibitem{geiger2013vision}
A.~Geiger, P.~Lenz, C.~Stiller, and R.~Urtasun, ``{Vision meets robotics: The
  KITTI dataset},'' \emph{The International Journal of Robotics Research},
  vol.~32, no.~11, pp. 1231--1237, 2013.

\bibitem{ros2016synthia}
G.~Ros, L.~Sellart \emph{et~al.}, ``{The SYNTHIA Dataset: A Large Collection of
  Synthetic Images for Semantic Segmentation of Urban Scenes},'' in \emph{IEEE
  Conference on Computer Vision and Pattern Recognition}, 2016, pp. 3234--3243.

\bibitem{Cordts2016Cityscapes}
M.~Cordts, M.~Omran \emph{et~al.}, ``{The Cityscapes Dataset for Semantic Urban
  Scene Understanding},'' in \emph{IEEE Conference on Computer Vision and
  Pattern Recognition (CVPR)}, 2016.

\bibitem{alhaija2018kittisemsegbenchmark}
H.~Alhaija, S.~Mustikovela \emph{et~al.}, ``Augmented reality meets computer
  vision: Efficient data generation for urban driving scenes,''
  \emph{International Journal of Computer Vision (IJCV)}, 2018.

\bibitem{yu2018bdd100k}
F.~Yu, W.~Xian \emph{et~al.}, ``{BDD100K: A Diverse Driving Video Database with
  Scalable Annotation Tooling},'' \emph{arXiv preprint arXiv:1805.04687}, 2018.

\bibitem{geyer2020a2d2}
\BIBentryALTinterwordspacing
J.~Geyer, Y.~Kassahun \emph{et~al.}, ``{A2D2: Audi Autonomous Driving
  Dataset},'' 2020. [Online]. Available: \url{https://www.a2d2.audi}
\BIBentrySTDinterwordspacing

\bibitem{Valada_2016_ISER}
A.~Valada, G.~Oliveira, T.~Brox, and W.~Burgard, ``{Deep Multispectral Semantic
  Scene Understanding of Forested Environments using Multimodal Fusion},'' in
  \emph{International Symposium on Experimental Robotics (ISER)}, Tokyo, Japan,
  Oct. 2016.

\bibitem{kragh2017fieldsafe}
M.~F. Kragh, P.~Christiansen \emph{et~al.}, ``{FieldSAFE: Dataset for Obstacle
  Detection in Agriculture},'' \emph{Sensors}, vol.~17, no.~11, p. 2579, 2017.

\bibitem{chebrolu2017agricultural}
N.~Chebrolu, P.~Lottes \emph{et~al.}, ``{Agricultural Robot Dataset for Plant
  Classification, Localization and Mapping on Sugar Beet Fields},'' \emph{The
  International Journal of Robotics Research}, vol.~36, no.~10, pp. 1045--1052,
  2017.

\bibitem{kame:geman1984stochastic}
S.~Geman and D.~Geman, ``{Stochastic relaxation, Gibbs distributions, and the
  Bayesian restoration of images},'' \emph{IEEE Transactions on Pattern
  Analysis and Machine Intelligence}, no.~6, pp. 721--741, 1984.

\bibitem{kame:kohli2008crf}
P.~{Kohli}, L.~{Ladicky}, and P.~H.~S. {Torr}, ``Robust higher order potentials
  for enforcing label consistency,'' in \emph{{Proc. IEEE Conf. Comput. Vision
  and Pattern Recognition (CVPR)}}, 2008, pp. 1--8.

\bibitem{kame:plath2009crf}
N.~Plath, M.~Toussaint, and S.~Nakajima, ``Multi-class image segmentation using
  conditional random fields and global classification,'' in \emph{Proceedings
  of the 26th Annual International Conference on Machine Learning}.\hskip 1em
  plus 0.5em minus 0.4em\relax Association for Computing Machinery, 2009, p.
  817–824.

\bibitem{kame:krizhevsky2012imagenet}
A.~Krizhevsky, I.~Sutskever, and G.~E. Hinton, ``{ImageNet Classification with
  Deep Convolutional Neural Networks},'' in \emph{Advances in Neural
  Information Processing Systems}, 2012, pp. 1097--1105.

\bibitem{kame:simonyan2015vgg}
K.~Simonyan and A.~Zisserman, ``Very deep convolutional networks for
  large-scale image recognition,'' in \emph{International Conference on
  Learning Representations}, 2015.

\bibitem{kame:long2015fully}
J.~Long, E.~Shelhamer, and T.~Darrell, ``{Fully Convolutional Networks for
  Semantic Segmentation},'' in \emph{IEEE Conference on Computer Vision and
  Pattern Recognition}, 2015, pp. 3431--3440.

\bibitem{kame:badrinarayanan2017segnet}
V.~Badrinarayanan, A.~Kendall, and R.~Cipolla, ``{SegNet: A Deep Convolutional
  Encoder-Decoder Architecture for Image Segmentation},'' \emph{IEEE
  Transactions on Pattern Analysis and Machine Intelligence}, vol.~39, no.~12,
  pp. 2481--2495, 2017.

\bibitem{kame:zhao2017pyramid}
H.~Zhao, J.~Shi \emph{et~al.}, ``{Pyramid Scene Parsing Network},'' in
  \emph{IEEE Conference on Computer Vision and Pattern Recognition}, 2017, pp.
  2881--2890.

\bibitem{kame:chen2017rethinking}
L.-C. Chen, G.~Papandreou, F.~Schroff, and H.~Adam, ``{Rethinking Atrous
  Convolution for Semantic Image Segmentation},'' \emph{arXiv preprint
  arXiv:1706.05587}, 2017.

\bibitem{kame:sun2019high}
K.~Sun, Y.~Zhao \emph{et~al.}, ``{High-Resolution Representations for Labeling
  Pixels and Regions},'' \emph{arXiv preprint arXiv:1904.04514}, 2019.

\bibitem{kame:paszke2016enet}
A.~Paszke, A.~Chaurasia, S.~Kim, and E.~Culurciello, ``{ENet: A Deep Neural
  Network Architecture for Real-Time Semantic Segmentation},'' \emph{arXiv
  preprint arXiv:1606.02147}, 2016.

\bibitem{kame:romera2017erfnet}
E.~Romera, J.~M. Alvarez, L.~M. Bergasa, and R.~Arroyo, ``{ERFNet: Efficient
  Residual Factorized ConvNet for Real-time Semantic Segmentation},''
  \emph{IEEE Transactions on Intelligent Transportation Systems}, vol.~19,
  no.~1, pp. 263--272, 2017.

\bibitem{kame:yu2018bisenet}
C.~Yu, J.~Wang \emph{et~al.}, ``{BiSeNet: Bilateral Segmentation Network for
  Real-time Semantic Segmentation},'' in \emph{European Conference on Computer
  Vision (ECCV)}, 2018, pp. 325--341.

\bibitem{kame:zhao2018icnet}
H.~Zhao, X.~Qi \emph{et~al.}, ``{ICNet for Real-Time Semantic Segmentation on
  High-Resolution Images},'' in \emph{European Conference on Computer Vision
  (ECCV)}, 2018, pp. 405--420.

\bibitem{kame:poudel2019fast}
R.~P. Poudel, S.~Liwicki, and R.~Cipolla, ``{Fast-SCNN: Fast Semantic
  Segmentation Network},'' \emph{arXiv preprint arXiv:1902.04502}, 2019.

\bibitem{kame:ji2020fgvc}
R.~Ji, L.~Wen \emph{et~al.}, ``{Attention Convolutional Binary Neural Tree for
  Fine-Grained Visual Categorization},'' in \emph{IEEE Conference on Computer
  Vision and Pattern Recognition (CVPR)}, June 2020.

\bibitem{kame:zheng2017fgvc}
H.~{Zheng}, J.~{Fu}, T.~{Mei}, and J.~{Luo}, ``{Learning Multi-attention
  Convolutional Neural Network for Fine-Grained Image Recognition},'' in
  \emph{IEEE International Conference on Computer Vision (ICCV)}, 2017, pp.
  5219--5227.

\bibitem{kame:kalogerakis2017shapepfcn}
E.~Kalogerakis, M.~Averkiou, S.~Maji, and S.~Chaudhuri, ``{3D Shape
  Segmentation with Projective Convolutional Networks},'' in \emph{IEEE
  Computer Vision and Pattern Recognition (CVPR)}, 2017.

\bibitem{kame:zhao20193dpointcapsule}
Y.~Zhao, T.~Birdal, H.~Deng, and F.~Tombari, ``{3D Point Capsule Networks},''
  in \emph{Conference on Computer Vision and Pattern Recognition (CVPR)}, 2019.

\bibitem{kame:liu2020partguidedediting}
Z.~Liu, F.~Lu \emph{et~al.}, ``{3D Part Guided Image Editing for Fine-Grained
  Object Understanding},'' in \emph{IEEE Conference on Computer Vision and
  Pattern Recognition (CVPR)}, June 2020.

\bibitem{kame:lu2018partstates}
C.~Lu, H.~Su \emph{et~al.}, ``{Beyond Holistic Object Recognition: Enriching
  Image Understanding With Part States},'' in \emph{IEEE Conference on Computer
  Vision and Pattern Recognition (CVPR)}, June 2018.

\bibitem{kame:lambert2020mseg}
J.~Lambert, Z.~Liu \emph{et~al.}, ``{MSeg: A Composite Dataset for Multi-Domain
  Semantic Segmentation},'' in \emph{IEEE Conference on Computer Vision and
  Pattern Recognition (CVPR)}, June 2020.

\bibitem{kame:himmelsbach2011autonomous}
M.~Himmelsbach, T.~Luettel \emph{et~al.}, ``{Autonomous Off-Road Navigation for
  MuCAR-3},'' \emph{KI-K{\"u}nstliche Intelligenz}, vol.~25, no.~2, pp.
  145--149, 2011.

\bibitem{tas:unterholzner2010iv-hybrid-adaptive-control}
A.~Unterholzner and H.-J. Wuensche, ``{Hybrid Adaptive Control of a Multi-Focal
  Vision System},'' in \emph{IEEE Intelligent Vehicles Symposium}, San Diego,
  CA, USA, Jun. 2010, pp. 534--539.

\bibitem{kame:russell2008labelme}
B.~C. Russell, A.~Torralba, K.~P. Murphy, and W.~T. Freeman, ``{LabelMe: A
  Database and Web-Based Tool for Image Annotation},'' \emph{International
  Journal of Computer Vision}, vol.~77, no. 1-3, pp. 157--173, 2008.

\bibitem{kame:yuan2018ocnet}
Y.~Yuhui and W.~Jingdong, ``{OCNet: Object Context Network for Scene
  Parsing},'' \emph{arXiv preprint arXiv:1809.00916}, 2018.

\bibitem{kame:fernandez2018new}
E.~Fernandez-Moral, R.~Martins, D.~Wolf, and P.~Rives, ``A new metric for
  evaluating semantic segmentation: leveraging global and contour accuracy,''
  in \emph{IEEE Intelligent Vehicles Symposium (IV)}.\hskip 1em plus 0.5em
  minus 0.4em\relax IEEE, 2018, pp. 1051--1056.

\bibitem{kame:paszke2019pytorch}
A.~Paszke, S.~Gross \emph{et~al.}, ``Pytorch: An imperative style,
  high-performance deep learning library,'' in \emph{Advances in Neural
  Information Processing Systems 32}, H.~Wallach, H.~Larochelle \emph{et~al.},
  Eds.\hskip 1em plus 0.5em minus 0.4em\relax Curran Associates, Inc., 2019,
  pp. 8024--8035.

\bibitem{kame:martin2015tensorflow}
M.~Abadi, A.~Agarwal \emph{et~al.}, ``{TensorFlow}: Large-scale machine
  learning on heterogeneous systems,'' 2015, software available from
  tensorflow.org.

\bibitem{nn:seif2018}
G.~Seif, ``{Semantic Segmentation Suite in TensorFlow},''
  \url{https://github.com/GeorgeSeif/Semantic-Segmentation-Suite}, 2018.

\bibitem{kame:Efficient-Segmentation-Networks}
Y.~Wang, ``{Efficient-Segmentation-Networks Pytorch Implementation},''
  \url{https://github.com/xiaoyufenfei/Efficient-Segmentation-Networks}, 2019.

\bibitem{kame:kingma2014adam}
D.~P. Kingma and J.~Ba, ``{ADAM: A method for stochastic optimization},''
  \emph{arXiv preprint arXiv:1412.6980}, 2014.

\bibitem{kame:he2016deep}
K.~He, X.~Zhang, S.~Ren, and J.~Sun, ``{Deep Residual Learning for Image
  Recognition},'' in \emph{IEEE Conference on Computer Vision and Pattern
  Recognition}, 2016, pp. 770--778.

\bibitem{kame:he2016identity}
------, ``{Identity Mappings in Deep Residual Networks},'' in \emph{European
  Conference on Computer Vision}.\hskip 1em plus 0.5em minus 0.4em\relax
  Springer, 2016, pp. 630--645.

\bibitem{kame:deng2009imagenet}
J.~Deng, W.~Dong \emph{et~al.}, ``{ImageNet: A Large-Scale Hierarchical Image
  Database},'' in \emph{{IEEE Conference on Computer Vision and Pattern
  Recognition}}.\hskip 1em plus 0.5em minus 0.4em\relax IEEE, 2009, pp.
  248--255.

\bibitem{kame:valada2017adapnet}
A.~Valada, J.~Vertens, A.~Dhall, and W.~Burgard, ``{AdapNet: Adaptive Semantic
  Segmentation in Adverse Environmental Conditions},'' in \emph{IEEE
  International Conference on Robotics and Automation (ICRA)}.\hskip 1em plus
  0.5em minus 0.4em\relax IEEE, 2017, pp. 4644--4651.

\bibitem{kame:treml2016speeding}
M.~Treml, J.~Arjona-Medina \emph{et~al.}, ``{Speeding up Semantic Segmentation
  for Autonomous Driving},'' in \emph{MLITS, NIPS Workshop}, vol.~2, no.~7,
  2016.

\bibitem{kame:chaurasia2017linknet}
A.~Chaurasia and E.~Culurciello, ``{LinkNet: Exploiting Encoder Representations
  for Efficient Semantic Segmentation},'' in \emph{IEEE Visual Communications
  and Image Processing (VCIP)}.\hskip 1em plus 0.5em minus 0.4em\relax IEEE,
  2017, pp. 1--4.

\bibitem{kame:casado2019clodsa}
{\'A}.~Casado-Garc{\'\i}a, C.~Dom{\'\i}nguez \emph{et~al.}, ``{CLoDSA: a tool
  for augmentation in classification, localization, detection, semantic
  segmentation and instance segmentation tasks},'' \emph{BMC Bioinformatics},
  vol.~20, no.~1, p. 323, 2019.

\bibitem{kame:shrivastava2016training}
A.~Shrivastava, A.~Gupta, and R.~Girshick, ``{Training Region-based Object
  Detectors with Online Hard Example Mining},'' in \emph{IEEE Conference on
  Computer Vision and Pattern Recognition}, 2016, pp. 761--769.

\end{thebibliography}

\end{document}